\documentclass{article}

\usepackage{microtype}
\usepackage{graphicx}
\usepackage{subcaption}
\usepackage{booktabs} % for professional tables

\usepackage{hyperref}

\usepackage[accepted]{icml2026}

\usepackage{amsmath}
\usepackage{amssymb}
\usepackage{mathtools}
\usepackage{amsthm}

\usepackage[capitalize,noabbrev]{cleveref}

\theoremstyle{plain}

\theoremstyle{definition}

\theoremstyle{remark}

\usepackage[textsize=tiny]{todonotes}

\usepackage{xspace}
\usepackage{multirow}
\usepackage{lipsum}
\usepackage{colortbl}
\usepackage{graphicx}
\usepackage{siunitx}
\usepackage{duckuments}
\usepackage{pifont} %xmark and cmark
\usepackage{bm}

\usepackage{algorithmic}

\newcommand{\algcomment}[1]{\hfill{\footnotesize$\triangleright$~#1}}

\definecolor{LightGray}{gray}{1}
\definecolor{lightgray}{gray}{0.95}
\definecolor{midgray}{gray}{0.55}
\definecolor{steelblue}{HTML}{4D82B7}
\definecolor{davysgrey}{rgb}{0.33, 0.33, 0.33}
\definecolor{LightCyan}{rgb}{0.88,1,1}
\definecolor{LightGold}{HTML}{F3E2C5}
\definecolor{AngelRow}{HTML}{FFFDD0}
\definecolor{ao(english)}{rgb}{0.0, 0.5, 0.0}
\definecolor{lightsalmon}{rgb}{1.0, 0.63, 0.48}
\definecolor{tabgreen}{HTML}{2CA02C}
\definecolor{newgreen}{HTML}{CCE0AC}
\definecolor{newblue}{HTML}{6bc5ff}
\newcommand{\ourcolor}{newblue!20}

\newcommand{\methodname}{{\textsc{Theseus}~}}

\newcommand{\red}[1]{\textcolor{red}{#1}}

\newcommand{\dgreen}[1]{\textcolor{ao(english)}{#1}}

\newcommand{\cmark}{\ding{51}}%
\newcommand{\greencmark}{\dgreen{\cmark}}
\newcommand{\xmark}{\ding{55}}%
\newcommand{\redxmark}{\red{\xmark}}

\newcommand{\tit}[1]{\smallbreak\noindent\textbf{#1 }}

\newcommand{\tinytit}[1]{\noindent\textbf{#1}}

\crefname{section}{Sec.}{Secs.}
\crefname{table}{Tab.}{Tabs.}
\crefname{figure}{Fig.}{Figs.}
\crefname{equation}{Eq.}{Eqs.}

\newcommand{\ourrow}{\rowcolor{\ourcolor}[\dimexpr\tabcolsep+0.1pt\relax]}

\makeatletter
\DeclareRobustCommand\onedot{\futurelet\@let@token\@onedot}
\def\@onedot{\ifx\@let@token.\else.\null\fi\xspace}

\def\eg{\emph{e.g}\onedot} 
\def\ie{\emph{i.e}\onedot}

\makeatother

\usepackage{array}
\newcommand{\PreserveBackslash}[1]{\let\temp=\\#1\let\\=\temp}
\newcolumntype{C}[1]{>{\PreserveBackslash\centering}p{#1}}
\newcolumntype{R}[1]{>{\PreserveBackslash\raggedleft}p{#1}}
\newcolumntype{L}[1]{>{\PreserveBackslash\raggedright}p{#1}}
\newcolumntype{Y}{>{\centering\arraybackslash}X}

\newcommand{\plus}{\texttt{+}}
\newcommand{\minus}{\texttt{-}}

\newcommand{\deltaneg}[1]{\textcolor{red}{\scriptsize{(\minus#1)}}}
\newcommand{\deltapos}[1]{\textcolor{green!60!black}{\scriptsize{(\plus#1)}}}
\newcommand{\deltazero}[0]{\textcolor{black}{\scriptsize{(\plus0.00)}}}
\newcommand{\deltanull}[0]{\textcolor{black}{\scriptsize{(-)}}}

\definecolor{MaterialRed}{HTML}{D32F2F}        % Red 700
\definecolor{MaterialBlue}{HTML}{1976D2}       % Blue 700
\definecolor{MaterialGreen}{HTML}{388E3C}      % Green 700
\definecolor{MaterialOrange}{HTML}{F57C00}     % Orange 700
\definecolor{MaterialPurple}{HTML}{7B1FA2}     % Purple 700
\definecolor{MaterialTeal}{HTML}{00796B}       % Teal 700
\definecolor{MaterialIndigo}{HTML}{303F9F}     % Indigo 700
\definecolor{MaterialBrown}{HTML}{5D4037}      % Brown 700

\icmltitlerunning{Transporting Task Vectors across Different Architectures without Training}

\begin{document}

\twocolumn[
  \icmltitle{Transporting Task Vectors across Different Architectures without Training}



  \icmlsetsymbol{equal}{*}

  \begin{icmlauthorlist}
    \icmlauthor{Filippo Rinaldi}{unimore}
    \icmlauthor{Aniello Panariello}{unimore}
    \icmlauthor{Giacomo Salici}{unimore}
    \icmlauthor{Angelo Porrello}{unimore}
    \icmlauthor{Simone Calderara}{unimore}
  \end{icmlauthorlist}

  \icmlaffiliation{unimore}{AImageLab, University of Modena and Reggio Emilia, Modena, Italy}

  \icmlcorrespondingauthor{Filippo Rinaldi}{filippo.rinaldi@unimore.it}

  \icmlkeywords{Machine Learning, ICML}

  \vskip 0.3in
]



\printAffiliationsAndNotice{}  % no special notice (required even if empty)
\begin{abstract}
Adapting large pre-trained models to downstream tasks often produces task-specific parameter updates that are expensive to relearn for every model variant. While recent work has shown that such updates can be transferred between models with identical architectures, transferring them across models of different widths remains unexplored. In this work, we introduce \textbf{\textsc{Theseus}}, a training-free method for transporting task updates across heterogeneous-width models. Rather than matching parameters, we characterize a task update by the functional effect it induces on intermediate representations. We formalize task-vector transport as a functional matching problem on observed activations and show that, after aligning representation spaces via orthogonal Procrustes analysis, it admits a stable closed-form solution that preserves the geometry of the update. We evaluate \methodname on vision and language models across different widths, showing consistent improvements over baselines without additional training or backpropagation. Our results show that task updates can be meaningfully transferred across architectures when task identity is defined functionally rather than parametrically. Code is available at \url{https://github.com/apanariello4/merge-and-rebase}.
\end{abstract}

\section{Introduction}
Large pre-trained models have become a cornerstone of modern machine learning, demonstrating remarkable performance across a wide range of tasks in computer vision \citep{dosovitskiy2021an,he2022masked} and natural language processing \citep{brown2020language,chowdhery2023palm}. A key factor contributing to their success is the ability to adapt these pre-trained models to downstream tasks through fine-tuning or other lightweight techniques \citep{houlsby2019parameter, liu2022few}. At the same time, practitioners often maintain families of models that share architectural similarities but differ in size, training data, or other characteristics, raising a natural question: \emph{can a task-specific update learned by one model be transferred directly to a different one without retraining?}
\begin{figure}[t]
    \centering
    \includegraphics[width=.97\linewidth]{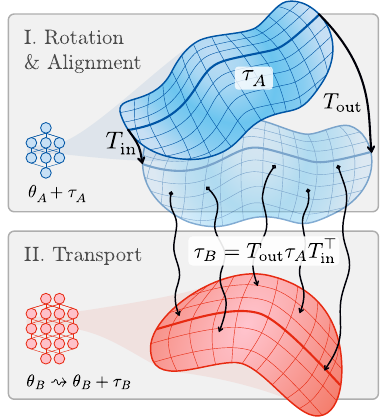}
    \vspace{-.4em}
    \caption{\textbf{Task-vector transport via orthogonal alignment.} Procrustes rotations identify shared structure between representation spaces, enabling training-free transfer of task updates across models with mismatched dimensions.}\label{fig:teaser}
    \vspace{-1em}
\end{figure}

Existing work on model transfer and rebasin has primarily focused on transferring knowledge between models of the same architecture and number of parameters \citep{ainsworth2023git,rinaldi2025update}. When models differ in size, however, the challenge becomes more pronounced due to the mismatch in parameter dimensions and representations. Previous approaches to this problem have relied on techniques such as knowledge distillation \citep{hinton2015distilling} or fine-tuning smaller models initialized from larger ones \citep{turc2019well,chen2022update} or vice versa \citep{chen2015net2net}. However, these methods usually require additional training and do not directly transfer the task-specific update itself.

In this work, we introduce a novel \emph{task-vector transport} framework for transferring task-specific parameter updates between models of different sizes (\cref{fig:teaser}). Rather than operating in parameter space, our approach focuses on how task updates modify intermediate representations, enabling transfer across models with mismatched widths or identical architectures.

Despite the apparent simplicity of transferring an update between two layers, this problem is fundamentally ill-posed when models differ in width. A task update learned in one model acts on a representation space whose geometry, dimensionality, and basis are generally incompatible with those of another model. Thus, directly resizing or projecting parameters does not preserve the functional effect of the update on the model's activations. Any meaningful transfer must therefore account for how task-specific updates interact with the internal representations induced by each model.

We address this challenge by formulating task-vector transport as a functional matching problem on observed activations. We characterize a task update by the bilinear form it induces between input and output activations of a layer, and seek a corresponding update in the target model that reproduces this effect. Since this problem is underdetermined across mismatched representation spaces, we resolve the ambiguity by aligning the representation spaces of the two models using orthogonal Procrustes analysis. This approach identifies shared subspaces that preserve norms and angles, yielding a closed-form, training-free transport rule.

At a conceptual level, this perspective resembles the classical Ship of Theseus~\citep{hobbes1656elements}. As model architectures evolve in width, depth, or pre-training data, their parameters may share no direct correspondence, yet we may still ask whether a task-specific update remains the same when transferred across models. We argue that task identity should be defined by functional behavior rather than parameter values. \methodname embodies this view by transporting task updates in a way that preserves their effect on representations. In this work, we focus primarily on transport across heterogeneous Transformer variants differing in width, depth, and pre-training distributions. In summary, our contributions are threefold:
\begin{itemize}
    \item We formulate task-vector transport across heterogeneous models as a functional matching problem based on the effect of task updates on layer activations.
    \item We introduce a Procrustes-based alignment that endows the transport problem with a principled geometric structure and a closed-form transport rule.
    \item We show that \methodname enables effective transfer of task-specific updates across models with different widths and pre-training distributions, outperforming vision and language baselines.
\end{itemize}

\section{Related Work}

\tit{Task updates and parameter-space transfer.}
Recent works have explored treating task-specific fine-tuning updates as first-class objects that can be manipulated, transferred, or composed in parameter space \citep{ilharco2023editing,marczak2025task,yadav2023tiesmerging,panariello2025modular,porrello2026dataless}. Rebasin methods exploit symmetries in neural networks to align weights across independently trained models with identical architectures, enabling interpolation or update transfer \citep{ainsworth2023git,singh2020fusion,rinaldi2025update}. Related work studies task arithmetic and model editing, where parameter differences encode task information that can be combined \citep{ilharco2023editing, wortsman2022model}.
Recent advances have introduced more sophisticated alignment strategies: \citet{nasery2025pleas} use permutations and least squares to optimize neuron alignment, while \citet{stoica2024zipit} enable the merging of models from distinct tasks by ``zipping'' similar features across layers. For transformer-based architectures, \citet{imfeld2024transformer} leverage optimal transport to reconcile the permutation symmetries inherent in attention blocks. However, these approaches rely on architectural equivalence and matching parameter dimensionality. When models differ in width or internal representation size, direct parameter alignment, padding, or projection no longer preserves the functional effect of the update, making the transfer problem ill-posed. Our work departs from parameter-centric alignment and instead defines task identity through its induced functional effect on representations, enabling transfer across models with mismatched widths.

\tit{Knowledge transfer and representation alignment.}
A large body of work addresses transfer across heterogeneous models using training-based techniques such as knowledge distillation, teacher--student learning, or architecture-aware transformations \citep{hinton2015distilling, turc2019well, chen2015net2net, kangaslahti2026boomerang}. While effective, these methods require additional optimization and relearn the task in the target model rather than transporting an existing update. Separately, representation alignment methods such as canonical correlation analysis and orthogonal Procrustes have been widely used to compare and translate neural representations across models, layers, and training runs \citep{panariello2025accurate, raghu2017svcca, tsv, williams2021generalized}. Related work has also explored transferring latent-space interventions and representation-space transformations across models \citep{moschella2023relative, maiorca2023latent}. In contrast, \methodname uses aligned representations to derive a closed-form transport rule for task-specific parameter updates through a functional operator-matching objective, enabling transport across models with mismatched dimensionalities. We provide a detailed empirical ablation comparing \methodname against feature translation methods in~\cref{app:feature_vs_operator}.

\section{Method}\label{sec:method}

\tinytit{Setting.} Let $\theta_A$ and $\theta_B$ denote the parameters of the base models $A$ and $B$, which share a common Transformer-style structure but may differ in width, depth, or pre-training. We treat $A$ as the source model and $B$ as the target model. Let $\ell$ denote a linear or attention submodule (\eg, an MLP projection or an attention head projection). After fine-tuning model $A$ on a downstream task, we define the updated weights of layer $\ell$ as $\theta_A^{\ell} + \tau_A^{\ell} \in \mathbb{R}^{d_{\mathrm{out},A} \times d_{\mathrm{in},A}}$. Our goal is to construct a corresponding update $\tau_B^{\ell} \in \mathbb{R}^{d_{\mathrm{out},B} \times d_{\mathrm{in},B}}$ for model $B$ such that the function implemented by layer $\ell$ after applying the update $\theta_B^{\ell} + \tau_B^{\ell}$ approximates the function implemented by the updated layer in model $A$.

Given a set of $N$ inputs we denote the input and output activations of layer $\ell$ in models $A$ and $B$ as:
\begin{equation}
\begin{aligned}
H_{\mathrm{in},A}^{(\ell)} &\in \mathbb{R}^{N \times L^A \times d_{\mathrm{in},A}}, \quad
H_{\mathrm{out},A}^{(\ell)} \in \mathbb{R}^{N \times L^A \times d_{\mathrm{out},A}},\\
H_{\mathrm{in},B}^{(\ell)} &\in \mathbb{R}^{N \times L^B \times d_{\mathrm{in},B}}, \quad
H_{\mathrm{out},B}^{(\ell)} \in \mathbb{R}^{N \times L^B \times d_{\mathrm{out},B}},
\end{aligned}
\end{equation}
where $L^A$ and $L^B$ are the sequence lengths of models $A$ and $B$, and $d_{\mathrm{in},A}, d_{\mathrm{out},A}, d_{\mathrm{in},B}, d_{\mathrm{out},B}$ are the input and output dimensions of the layer $\ell$ in models $A$ and $B$ respectively. In the following, we drop the superscript $\ell$ to ease the notation.

\tit{Token Preprocessing.} To align the input activations of the two models, we first ensure that they have the same sequence length. If $L^A < L^B$, we interpolate the sequence length of model $A$ to match $L^B$ using bilinear interpolation. We then flatten the first two dimensions of the input activations; let $M = N \times L^B$ be the total number of tokens after interpolation. The input activations can then be represented as $H_{\mathrm{in},A} \in \mathbb{R}^{M \times d_{\mathrm{in},A}}$ and $H_{\mathrm{in},B} \in \mathbb{R}^{M \times d_{\mathrm{in},B}}$, and similarly for the output activations. We evaluate different sequence alignment strategies in \cref{sec:interpolation_strategies}.

\tit{Objective.} We recast task-vector transport as the problem of preserving the \emph{functional effect} of a task update on observed activations. Consider a linear layer whose output activations are given by $H_{\mathrm{out}} = H_{\mathrm{in}} \theta^\top$, where $\theta$ denotes the layer weight matrix of the base model. Applying a task update $\tau_A$ to model $A$ induces a change in the output activations
\begin{equation}
\label{eq:delta_h_out}
\Delta H_{\mathrm{out},A} = H_{\mathrm{in},A} \tau_A^\top .
\end{equation}
This expression characterizes how the update modifies the layer outputs for the given inputs. Rather than matching $\Delta H_{\mathrm{out},A}$ pointwise, we characterize the update by how it \emph{couples} input and output activations across the dataset. Specifically, we define the matrix
\begin{equation}
\label{eq:target_bilinear}
G_A \;=\; \Delta H_{\mathrm{out},A}  H_{\mathrm{out},A}^\top \in \mathbb{R}^{M \times M},
\end{equation} which aggregates the interaction between input activations, the task update, and the pre-update output representations. For any pair of tokens $(i,j)$, the entry ${(G_A)}_{ij}$ measures how the update-induced change at token $i$ aligns with the original output representation at token $j$. $G_A$ defines a bilinear form over token representations that captures the global functional signature of the task update on the observed data.

Thus, we seek an update $\tau_B$ for model $B$ whose induced bilinear form matches that of $\tau_A$:
\begin{equation}
\label{eq:transport_obj}
\min_{\tau_B}\;
{\left\|
H_{\mathrm{in},A}\tau_A^{\top}H_{\mathrm{out},A}^{\top}
-
H_{\mathrm{in},B}\tau_B^{\top}H_{\mathrm{out},B}^{\top}
\right\|}_F^2\ .
\end{equation}
Matching these bilinear forms ensures that $\tau_B$ reproduces the same input-output interaction structure as $\tau_A$, even when the two models operate in different representation spaces.

While~\cref{eq:transport_obj} provides a principled formulation of task-vector transport, it admits many solutions when the representation spaces of the two models are mismatched or only partially observed. In particular, unconstrained least-squares alignment becomes underdetermined whenever the sampled activations are rank-deficient, allowing solutions that arbitrarily amplify poorly conditioned directions. Rather than resolving this ambiguity through direct matrix inversion, we impose orthogonality constraints on the alignment maps. This removes scaling ambiguities, regularizes the transport problem, and preserves the norm and geometry of the transported update. \Cref{app:pinv_relation} provides the full derivation showing how the constrained problem reduces to a trace maximization solved in closed form by orthogonal Procrustes.
\begin{algorithm}[t]
    \caption{Training-free Task-Vector Transport}\label{theseus_alg}
    \begin{algorithmic}[1]
        \REQUIRE Source weights $\theta_A, \theta_A^{ft}$, target weights $\theta_B$, calibration dataset $\mathcal{D}$, scaling $\alpha$, sequence lengths $L^A, L^B$.
        \FOR{each layer $l \in \{1, \dots, L\}$}
            \STATE Compute source task vector: $\tau_A^l \leftarrow \theta_A^{ft,l} - \theta_A^l$
            \STATE Collect ${\{H_{in,h}, H_{out,h}\}}_{h \in \{A,B\}}$ from $\mathcal{D}$
            \STATE \textbf{if} $L^A \neq L^B$ \textbf{then} interpolate $\min\{L^A,L^B\}$
            \STATE Compute cross-covariances: \\
             \hspace{1em} $C_{in} \leftarrow H_{in,A}^\top H_{in,B},\quad  C_{out} \leftarrow H_{out,A}^\top H_{out,B}$
            \STATE Compute SVD: \algcomment{\cref{eq:svd_crosscovariances}}\\
             \hspace{1em} $C_{in} = U_{in}\Sigma_{in}V_{in}^\top$, $C_{out} = U_{out}\Sigma_{out}V_{out}^\top$
            \STATE Solve orthogonal Procrustes problems: \algcomment{\cref{eq:procrustes}}\\
            \hspace{1em} $T_{in} \leftarrow U_{in}V_{in}^\top,\quad T_{out} \leftarrow U_{out}V_{out}^\top$
            \STATE Compute transported update: \algcomment{\cref{eq:tauB_solution}}\\
            \hspace{1em} $\tau_B^l \leftarrow T_{out}\tau_A^l T_{in}^\top$
        \ENDFOR
        \STATE Return $\theta_B +\alpha \tau_B$
    \end{algorithmic}
\end{algorithm}

\tit{Geometric Alignment via Procrustes.}
We therefore align the representation spaces of the two models through orthonormal embeddings. Concretely, we seek orthogonal alignment maps that identify the maximally aligned shared subspaces while preventing arbitrary rescaling of poorly conditioned directions. This alignment is obtained by solving two orthogonal Procrustes problems:
\begin{equation}
\label{eq:procrustes_obj}
\begin{aligned}
    \min_{T_{\mathrm{in}}} \ &\| H_{\mathrm{in},A} T_{\mathrm{in}} - H_{\mathrm{in},B} \|_F^2
    \quad \text{s.t.}  \quad T_{\mathrm{in}}T_{\mathrm{in}}^\top = I_{d_{\mathrm{in},A}} ,\\
    \min_{T_{\mathrm{out}}}\  &\| H_{\mathrm{out},A} T_{\mathrm{out}} - H_{\mathrm{out},B} \|_F^2
    \quad \text{s.t.} \quad T_{\mathrm{out}}T_{\mathrm{out}}^\top = I_{d_{\mathrm{out},A}}.
\end{aligned}
\end{equation}
Here $T_{\mathrm{in}} \in \mathbb{R}^{d_{\mathrm{in},A} \times d_{\mathrm{in},B}}$ and
$T_{\mathrm{out}} \in \mathbb{R}^{d_{\mathrm{out},A} \times d_{\mathrm{out},B}}$
are rectangular matrices with orthonormal rows. They define orthogonal alignments between the dominant shared subspaces of the two models, while accommodating mismatched widths without requiring a bijective feature correspondence. Both admit closed-form solutions via Singular Value Decomposition (SVD) of the cross-covariances $C_{\mathrm{in}} = H_{\mathrm{in},A}^\top H_{\mathrm{in},B}$, and $C_{\mathrm{out}} = H_{\mathrm{out},A}^\top H_{\mathrm{out},B}$:
\begin{equation}\label{eq:svd_crosscovariances}
    C_{\mathrm{in}} = U_{\mathrm{in}} \Sigma_{\mathrm{in}} V_{\mathrm{in}}^\top,
    \quad C_{\mathrm{out}} = U_{\mathrm{out}} \Sigma_{\mathrm{out}} V_{\mathrm{out}}^\top,
\end{equation}
yielding
\begin{equation}\label{eq:procrustes}
    T_{\mathrm{in}} = U_{\mathrm{in}} V_{\mathrm{in}}^\top,
    \quad T_{\mathrm{out}} = U_{\mathrm{out}} V_{\mathrm{out}}^\top.
\end{equation}
The resulting embeddings minimize the discrepancy between aligned activations in a least-squares sense. Consequently, the activations of model $B$ can be approximated through aligned representations of model $A$:
\begin{equation}
\label{eq:procrustes_approx}
    H_{\mathrm{in},B} \approx H_{\mathrm{in},A} T_{\mathrm{in}},
    \qquad
    H_{\mathrm{out},B} \approx H_{\mathrm{out},A} T_{\mathrm{out}}.
\end{equation}
When the target dimensionality exceeds the source dimensionality, the transport acts as a norm-preserving embedding into a higher-dimensional aligned subspace. Conversely, when the target dimensionality is smaller, it reduces to a projection onto the maximally aligned shared subspace. Thus, the approximation in~\cref{eq:procrustes_approx} reflects the fact that transport is restricted to the shared aligned components of the two representation spaces. Substituting these aligned representations into~\cref{eq:transport_obj}, we obtain the following \emph{surrogate} objective:
\begin{equation*}
\label{eq:aligned_transport_obj}
\min_{\tau_B}
\left\|
H_{\mathrm{in},A}\tau_A^{\top}H_{\mathrm{out},A}^{\top}
-
(H_{\mathrm{in},A} T_{\mathrm{in}}) \tau_B^{\top} {(H_{\mathrm{out},A} T_{\mathrm{out}})}^{\top}
\right\|_F^2
\end{equation*}
which can be rewritten as
\begin{equation}
\label{eq:aligned_transport_obj_inner}
\min_{\tau_B}\;
\big\|
H_{\mathrm{in},A}\,E\,H_{\mathrm{out},A}^{\top}
\big\|_F^2,
\quad
E := \tau_A^{\top} - T_{\mathrm{in}} \tau_B^{\top} T_{\mathrm{out}}^{\top}.
\end{equation}
This formulation restricts transport to the aligned shared subspaces identified by Procrustes. Under this aligned-subspace approximation, the transport objective is minimized when the residual term vanishes, \ie, $E = 0$, yielding the closed-form transport rule
\begin{equation}
\label{eq:tauB_solution}
    \tau_B = T_{\mathrm{out}} \tau_A T_{\mathrm{in}}^{\top}.
\end{equation}

Thus, the transported update $\tau_B$ is obtained by orthogonally transporting the source update $\tau_A$ into the target representation space while preserving its dominant functional effect under the aligned-subspace approximation. A full derivation is provided in~\cref{app:closed_form_derivation}. We summarize the complete pipeline in~\cref{theseus_alg}.

\subsection{Properties and Scope}
\label{sec:norm_preservation}
\tit{\methodname Preserves the Norm.} A key property of the transport rule in~\cref{eq:tauB_solution} is that it preserves the Frobenius norm of the task update:
\begin{equation}
\label{eq:norm_preservation}
\|\tau_B\|_F = \|T_{\mathrm{out}} \tau_A T_{\mathrm{in}}^{\top}\|_F = \|\tau_A\|_F,
\end{equation}
since $T_{\mathrm{in}}$ and $T_{\mathrm{out}}$ have orthonormal rows. This norm preservation ensures that the magnitude of the update remains consistent across models, preventing unintended scaling effects and preserving the relative strength of the task update across architectures. Beyond preserving the update magnitude, orthogonality also stabilizes the transport problem. Indeed, in unconstrained least-squares formulations, rank-deficient activation covariances admit infinitely many equivalent solutions that can arbitrarily amplify poorly conditioned directions. By restricting the transport maps to orthogonal transformations, we remove these scaling ambiguities while preserving the norm and relative geometry of the update within the aligned subspace.

\tit{On the Generalization of \textsc{Theseus}.} While we present the method assuming model $A$ has lower-dimensional representations than model $B$, the formulation is symmetric. In practice, task-vector transport can be applied in either direction by swapping the roles of the two models. Moreover, when the two models have representations of equal dimensionality, the Procrustes mappings reduce to square orthogonal transformations, and the same transport rule applies without modification. We empirically study both settings in \cref{sec:experiments}.

\tit{Extension to Different Depths.} Our formulation assumes that the two models have the same number of layers; however, it can be extended to different depths by combining \methodname with a separate layer-matching or interpolation strategy. We provide preliminary results on this extension in \cref{sec:transport_diff_arch}, leaving a detailed exploration to future work.

\section{Experiments}\label{sec:experiments}
\begin{table*}[t]
\def\arraystretch{1.04}
  \caption{\textbf{Width Scaling (Narrow $\to$ Wide).} Task-vector transport from a ViT-B/16 model pre-trained on \texttt{LAION-2B} ($A$) to the wider ViT-B/16+ model pre-trained on \texttt{LAION-400M} ($B$). Results are reported for different numbers of alignment batches $\mathcal{B}$ used to estimate the Procrustes maps. $\Delta$Acc denotes the accuracy improvement over the zero-shot baseline of model $B$. \methodname represents $\theta_B + \tau_B$.}
  \label{tab:small_to_big_batch}
  \centering
\resizebox{\linewidth}{!}{
\begin{tabular}{l c c c c c c c c c c}
\toprule
\textbf{Model} & $\mathcal{B}$ & \textbf{EUROSAT} & \textbf{SVHN} & \textbf{GTSRB} & \textbf{RESISC45} & \textbf{DTD} & \textbf{CARS} & \textbf{MNIST} & \textbf{SUN397} & \textbf{AVG ($\Delta$Acc)} \\
\midrule
$\theta_{B}$ \textit{zero-shot}
& \multicolumn{1}{c}{--} & 50.92 & 39.23 & 49.63 & 64.53 & 55.48 & 84.53 & 57.06 & 68.67 & 58.76 \deltazero \\

$\theta_{B}$ \textit{fine-tune}
& \multicolumn{1}{c}{--} & 98.96 & 91.08 & 98.63 & 92.59 & 77.81 & 87.65 & 99.63 & 76.76 & 90.39 \deltapos{31.63} \\

$\theta_{A}$ \textit{fine-tune}
& \multicolumn{1}{c}{--} & 98.69 & 97.45 & 98.64 & 95.65 & 82.24 & 91.53 & 99.61 & 79.86 & 92.96 \deltapos{34.20} \\

\midrule

$\theta_{B} + \tau_{A}^{\text{padded}}$
& \multicolumn{1}{c}{--} & 48.82 & 38.17 & 48.45 & 65.14 & 54.94 & 84.52 & 58.52 & 69.46 & 58.50 \deltaneg{0.26} \\

$\theta_{B} + \tau_{B}^{\text{random}}$
& \multicolumn{1}{c}{--} & 50.63 & 38.96 & 49.65 & 65.65 & 55.58 & 84.56 & 57.39 & 69.60 & 59.00 \deltapos{0.24} \\

\midrule

$\theta_{B} + \tau_{\text{pinv}}$
& 1 & 14.29 & 14.54 & 31.66 & 54.73 & 48.94 & 69.61 & 40.49 & 64.79 & 42.38 \deltaneg{16.38} \\

$\theta_{B} + \tau_{\text{pinv-Tikh}}$
& 1 & 47.83 & 38.24 & 50.23 & 63.55 & 56.12 & 82.29 & 54.68 & 68.49 & 57.68 \deltaneg{1.08} \\

$\theta_{B} + \tau_{A}^{\text{random}\rightarrow B}$
& 1 & 51.53 & 39.49 & 49.79 & 65.44 & 55.53 & 84.31 & 57.33 & 69.61 & 59.13 \deltapos{0.37} \\
\ourrow
\methodname
& 1 & 60.41 & 56.80 & 56.62 & 66.98 & 57.23 & 84.62 & 64.49 & 69.60 & \textbf{64.59 \deltapos{5.83}} \\

\midrule

$\theta_{B} + \tau_{\text{pinv}}$
& 2 & 44.34 & 41.92 & 48.19 & 64.22 & 56.43 & 77.59 & 20.44 & 68.88 & 52.25 \deltaneg{6.51} \\

$\theta_{B} + \tau_{\text{pinv-Tikh}}$
& 2 & 53.04 & 46.23 & 50.11 & 64.52 & 57.18 & 81.83 & 55.13 & 68.76 & 59.60 \deltapos{0.84} \\

$\theta_{B} + \tau_{A}^{\text{random}\rightarrow B}$
& 2 & 50.96 & 39.41 & 49.75 & 65.60 & 55.48 & 84.41 & 56.62 & 69.51 & 58.97 \deltapos{0.21} \\
\ourrow
\methodname
& 2 & 64.03 & 59.30 & 59.55 & 68.12 & 57.86 & 84.89 & 68.97 & 70.12 & \textbf{66.60 \deltapos{7.84}} \\

\midrule

$\theta_{B} + \tau_{\text{pinv}}$
& 5 & 52.25 & 44.82 & 50.12 & 64.24 & 56.92 & 82.51 & 27.72 & 69.85 & 56.80 \deltaneg{1.96} \\

$\theta_{B} + \tau_{\text{pinv-Tikh}}$
& 5 & 54.51 & 46.16 & 50.34 & 65.53 & 57.93 & 82.12 & 56.97 & 68.76 & 60.29 \deltapos{1.53} \\

$\theta_{B} + \tau_{A}^{\text{random}\rightarrow B}$
& 5 & 50.79 & 39.24 & 49.69 & 65.54 & 55.37 & 84.26 & 56.92 & 69.55 & 58.92 \deltapos{0.16} \\

\ourrow
\methodname
& 5 & 63.59 & 59.61 & 60.46 & 68.40 & 58.83 & 84.85 & 67.01 & 70.35 & \textbf{66.64 \deltapos{7.88}} \\

\midrule
\ourrow
\methodname
& 10 & 65.91 & 61.59 & 61.23 & 68.73 & 59.52 & 84.93 & 74.46 & 70.42 & \textbf{68.35 \deltapos{9.59}} \\

\midrule
\ourrow
\methodname
& 20 & 67.01 & 62.52 & 61.01 & 69.27 & 60.48 & 85.04 & 76.66 & 70.65 & \textbf{69.08 \deltapos{10.32}} \\
\midrule
\ourrow
\methodname
& 50 & 67.13 & 64.29 & 59.46 & 70.01 & 61.28 & 85.19 & 77.23 & 71.28 & \textbf{69.48 \deltapos{10.72}}\\
\bottomrule
\end{tabular}
  }
\end{table*}
\vspace{1em}

We evaluate the efficacy of our transport method across diverse vision and language benchmarks. Our experimental evaluation is organized around three questions. First, we assess our training-free functional matching approach when transporting updates between models of \textbf{different widths}. Second, we evaluate transfer across different pre-training distributions for models with \textbf{identical architectures}, enabling direct comparison with prior rebasin methods. Finally, we move beyond the core assumptions of our framework and evaluate a challenging regime in which both depth and width differ (\eg, ViT-B $\rightarrow$ ViT-L), showing that \methodname enables transport across \textbf{different widths and depths}.

\subsection{Experimental Setting}
Task-specific updates are obtained by fine-tuning a source model on a downstream task and are transported to a target model without additional optimization. \methodname is applied to all Transformer linear submodules, including attention projections and MLP layers, while positional and patch embeddings are excluded. Alignment operations are performed using frozen models and require only forward passes. Full implementation details are provided in \cref{sec:exp_setting}.

\tit{Activation Collection.} Procrustes maps ($T_{\mathrm{in}}$ and $T_{\mathrm{out}}$) are estimated independently for each layer using activations collected from calibration batches. For vision models, all spatial tokens are retained after sequence interpolation, while for language models we use the full token sequence. A naive computation would require storing the full activation tensors $H_{\mathrm{in},A}, H_{\mathrm{in},B}, H_{\mathrm{out},A}, H_{\mathrm{out},B}$, each of shape $\mathbb{R}^{M \times d}$, in memory. Under this approach, the memory footprint would grow as the total token count $M$ increases with every additional calibration batch. Instead, we compute activation statistics incrementally by aggregating the cross-covariance matrices
$C_{\mathrm{in}} = H_{\mathrm{in},A}^{\top}H_{\mathrm{in},B} \in \mathbb{R}^{d_{\mathrm{in},A} \times d_{\mathrm{in},B}}$
and $C_{\mathrm{out}} = H_{\mathrm{out},A}^{\top}H_{\mathrm{out},B} \in \mathbb{R}^{d_{\mathrm{out},A} \times d_{\mathrm{out},B}}$ across batches, keeping the memory footprint bounded by the models' internal feature dimensions.

This implementation decouples the memory requirements from the number of calibration tokens $M$, allowing the alignment process to scale to large calibration sets. The resulting covariance estimates are then used to compute the orthogonal Procrustes alignments via SVD. Further details and complexity analysis are provided in \cref{sec:comp_efficiency_suppl}.

\tit{Few-shot protocol.}
We evaluate our method using a $\mathcal{B}$-shot protocol, where $\mathcal{B} \in \{1, 2, 5, 10, 20\}$ denotes the number of sampled batches, each with a fixed batch size of $32$, used to characterize the representation spaces of models $A$ and $B$. For each value of $\mathcal{B}$, we sample training batches uniformly at random to construct the support set required for Procrustes alignment (\cref{eq:procrustes_obj}). Additionally, to maintain consistency with existing literature, we also report results under a standard $\mathcal{K}$-shot evaluation protocol with $\mathcal{K} \in \{1, 2, 5, 10, 20\}$ examples per class.

\tit{Update Scaling.}
Following standard practice in the task-vector and model editing literature~\citep{ilharco2023editing,tsv,marczak2025task,yadav2023tiesmerging,wortsman2022robust,sommariva2026distilling}, we integrate the transported update into the base model through a scalar interpolation $\alpha$, chosen via a validation set, applied as $\theta + \alpha \tau$. All reported results correspond to the value of $\alpha$ that achieves the best performance in a linear search.

\subsection{Transfer across Mismatched Widths}\label{sec:transport_width}
\tinytit{Vision.} We evaluate task-vector transport across models of differing widths in a vision setting. Specifically, we transport task updates $\tau_A$ learned by a CLIP ViT-B/16 model pre-trained on \texttt{LAION-2B} to the wider ViT-B/16+ variant from OpenCLIP~\citep{cherti2023reproducible}, pre-trained on \texttt{LAION-400M}. We also evaluate reverse transport from the wider model to a smaller ViT-B/16 target ($B \to A$). In this reverse setting, the target model is a ViT-B/16 pre-trained on the stronger \texttt{DataComp-XL} dataset. We evaluate on the 8-Vision benchmark~\citep{ilharco2023editing}; more details are provided in \cref{sec:vision_exp}.

\tit{Baselines.}
For width upscaling transfer, we compare our method against several baselines in \cref{tab:small_to_big_batch}. We report the zero-shot performance of the target model $B$ as a lower bound and fully fine-tuned models $A$ and $B$ as upper bounds. As naive baselines, we include: (i) zero-padding the source update $\tau_A$ to match the dimensionality of model $B$, and (ii) adding a randomly initialized update to model $B$. Both ignore representation alignment and test whether dimensional compatibility alone preserves task behavior. We also evaluate a pseudo-inverse transport baseline ($\tau_{\text{pinv}}$), which solves \cref{eq:transport_obj} using the Moore--Penrose pseudo-inverse of the activation matrices, implemented through the truncated SVD approximation in \texttt{torch.linalg.pinv}.

We further consider a Tikhonov-regularized variant ($\tau_{\text{pinv-Tikh}}$). While regularization improves numerical stability, both methods lack explicit representation alignment. Additional ablations are reported in~\cref{app:feature_vs_operator}. Finally, we include a randomized alignment baseline ($\tau_A^{\text{random}\rightarrow B}$), where a random update is transported through the learned Procrustes maps.

\tit{Results.}
As shown in~\cref{tab:small_to_big_batch}, \methodname consistently improves over the zero-shot baseline for forward transfer ($A \to B$) across all $\mathcal{B}$-shot regimes, without requiring gradient-based updates on the target model. While a gap to supervised fine-tuning remains, the method recovers a substantial fraction of the achievable gains, particularly in low-shot settings. Additional results for the $\mathcal{K}$-shot protocol and reverse transport ($B \to A$) are reported in \cref{sec:width_additional}, where we observe similar trends.

In the reverse setting, the target model $A$ is pre-trained on \texttt{DataComp-XL}, yielding a stronger zero-shot baseline. Nevertheless, \methodname consistently improves performance even in this high-accuracy regime, showing that transported updates can inject task-specific knowledge into already strong base models. In both transfer directions, the proposed alignment is substantially more stable than direct matrix-inversion approaches, whose pseudo-inverse variants deteriorate in low-shot regimes due to rank-deficient activation matrices.% Even when augmented with norm-based regularization, the pseudo-inverse approach remains less stable and consistently underperforms our closed-form transport rule, highlighting the benefits of explicitly identifying shared representation subspaces via orthogonal alignment.

\begin{table}[t]
\def\arraystretch{1.1}
  \caption{Linear-probing performance with and without the transported task vector. Source: \texttt{T5-3B} ($A$), target: \texttt{T5-Large} ($B$). \methodname represents $\theta_B + \tau_B$.}
  \label{tab:t5_lp}
  \vspace{.5em}
  \centering
  \setlength{\tabcolsep}{4pt}
  \resizebox{\linewidth}{!}{
  \begin{tabular}{l c c c c c c c}
    \toprule
    \textbf{Encoder} & $\mathcal{B}$ & \textbf{MNLI} & \textbf{QNLI} & \textbf{RTE} & \textbf{SCITAIL} & \textbf{SNLI} & \textbf{AVG ($\Delta$Acc)} \\
    \midrule
    $\theta_A$ \small{\textit{fine-tune}}& All & 89.88 & 95.78 & 87.73 & 92.45 & 90.45 & 91.26 \deltanull \\
    $\theta_B$ \small{\textit{fine-tune}}& All & 86.34 & 92.14 & 81.23 & 91.12 & 88.75 & 87.92 \deltanull \\
    \midrule
    $\theta_A$ & 20 & 66.84 & 80.85 & 71.84 & 69.40 & 45.06 & 66.80 \deltapos{8.39} \\
    $\theta_B$ & 20 & 64.87 & 66.61 & 64.62 & 49.85 & 43.42 & 57.87 \deltazero \\
    \ourrow
    \methodname & 20 & 78.67 & 86.09 & 74.00 & 84.74 & 76.04 & \textbf{79.91 \deltapos{22.04}} \\
    \midrule
    $\theta_A$ & 50 & 74.91 & 82.43 & 74.01 & 68.67 & 72.61 & 74.53     \deltaneg{2.25} \\
    $\theta_B$ & 50 & 78.58 & 81.79 & 76.17 & 75.15 & 72.20 & 76.78 \deltazero \\
    \ourrow
    \methodname & 50 & 81.81 & 89.10 & 80.90 & 88.30 & 80.30 & \textbf{84.08 \deltapos{7.30} }\\
    \midrule
    $\theta_A$ & 100 & 80.14 & 90.59 & 79.42 & 82.06 & 80.78 & 82.60 \deltapos{0.69} \\
    $\theta_B$ & 100 & 81.04 & 84.55 & 81.59 & 82.36 & 80.02 & 81.91 \deltazero \\
    \ourrow
    \methodname & 100 & 83.60 & 90.70 & 83.80 & 89.00 & 82.10 & \textbf{85.84 \deltapos{3.93}} \\
    \bottomrule
  \end{tabular}}
\end{table}

\begin{table*}[t]
\def\arraystretch{1.04}
\caption{\textbf{Identical Architecture Transfer.}
We report task-vector transport accuracies between ViT-B/16 models with identical architectures but different pre-training distributions ($A$: \texttt{DataComp-XL} $\to$ $B$: \texttt{LAION-2B}). Results are reported for varying support set sizes $\mathcal{K}$ used for alignment or optimization. $\Delta$Acc denotes the average accuracy improvement over the zero-shot baseline of model $B$. The \textbf{No Grad.} column indicates whether the method requires gradients. \methodname represents $\theta_B + \tau_B$.}\label{tab:same_arch}
    \centering
  \resizebox{\linewidth}{!}{
  \begin{tabular}{lcC{1.38cm} *{9}{c}}
    \toprule
    \textbf{Model} & $\mathcal{K}$ & \textbf{No Grad.} &
    \textbf{EUROSAT} & \textbf{SVHN} & \textbf{GTSRB} & \textbf{RESISC45} & \textbf{DTD} &
    \textbf{CARS} & \textbf{MNIST} & \textbf{SUN397} & \textbf{AVG ($\Delta$Acc)} \\
    \midrule
    $\theta_{B}$ \textit{zero-shot} & -- & \greencmark
      & 49.41 & 50.58 & 48.29 & 67.98 & 55.96 & 84.50 & 57.10 & 68.70 & 60.31 \deltazero \\
    $\theta_{B}$ \textit{fine-tune} & -- & \redxmark
      & 98.70 & 97.45 & 98.65 & 95.66 & 83.19 & 88.10 & 99.60 & 77.20 & 92.32 \deltapos{32.01} \\
    $\theta_{B} + \tau_A$ & -- & \greencmark
      & 49.58 & 50.84 & 49.31 & 67.87 & 56.27 & 84.60 & 57.40 & 68.90 & 60.60 \deltapos{0.28} \\
    TransFusion & -- & \greencmark 
      & 50.12 & 67.99 & 50.24 & 53.26 & 56.70 & 84.96 & 59.26 & 69.12 & 61.46 \deltapos{1.14} \\
    \midrule
    GradFix & 1 & \redxmark
      & 61.94 & 71.07 & 60.88 & 70.05 & 58.32 & 87.95 & 86.39 & 72.67 & 71.16 \deltapos{10.84} \\
    \ourrow
    \methodname & 1 & \greencmark
      & 58.64 & 71.38 & 60.30 & 71.95 & 59.79 & 88.89 & 87.17 & 73.08 & \textbf{71.40 \deltapos{11.09}} \\
      \ourrow
      \quad + GradFix & 1 & \redxmark
      & 62.09 & 73.04 & 64.13 & 70.66 & 58.62 & 87.92 & 88.15 & 72.93 & \textbf{72.19 \deltapos{11.88}} \\
    \midrule
    GradFix & 2 & \redxmark
      & 65.07 & 70.19 & 64.33 & 71.42 & 58.51 & 87.15 & 88.38 & 72.92 & 72.25 \deltapos{11.93} \\
      \ourrow
    \methodname & 2 & \greencmark
      & 65.49 & 75.66 & 62.67 & 72.38 & 60.48 & 88.85 & 91.56 & 73.02 & \textbf{73.76 \deltapos{13.45}} \\
      \ourrow
      \quad + GradFix & 2 & \redxmark
      & 67.73 & 75.53 & 66.51 & 72.49 & 60.21 & 87.18 & 90.67 & 73.14 & \textbf{74.18 \deltapos{13.87}} \\
    \midrule
    GradFix & 5 & \redxmark
      & 66.05 & 73.59 & 66.61 & 71.57 & 60.02 & 86.82 & 89.06 & 72.97 & 73.34 \deltapos{13.02} \\
    \ourrow
    \methodname & 5 & \greencmark
      & 66.12 & 72.09 & 61.82 & 73.33 & 61.48 & 88.80 & 93.03 & 73.19 & \textbf{73.73 \deltapos{13.42}} \\
      \ourrow
      \quad + GradFix & 5 & \redxmark
      & 69.48 & 74.72 & 68.48 & 72.55 & 61.22 & 87.01 & 90.16 & 73.16 & \textbf{74.60 \deltapos{14.28}} \\
    \midrule
    GradFix & 10 & \redxmark
      & 66.59 & 74.82 & 66.02 & 72.05 & 60.18 & 86.64 & 89.67 & 73.18 & 73.64 \deltapos{13.33} \\
    \ourrow
    \methodname & 10 & \greencmark
      & 65.21 & 73.95 & 63.46 & 74.09 & 62.29 & 88.83 & 92.48 & 73.21 & \textbf{74.19 \deltapos{13.88}} \\
      \ourrow
      \quad + GradFix & 10 & \redxmark
      & 67.63 & 76.92 & 68.11 & 72.62 & 61.44 & 86.29 & 90.68 & 73.33 & \textbf{74.63 \deltapos{14.31}} \\
    \midrule
    GradFix & 20 & \redxmark
      & 67.05 & 74.11 & 66.42 & 72.29 & 60.92 & 85.99 & 89.63 & 73.06 & 73.68 \deltapos{13.37} \\
    \ourrow
    \methodname & 20 & \greencmark
      & 68.42 & 74.82 & 65.73 & 74.16 & 63.77 & 88.92 & 94.62 & 73.33 & \textbf{75.47 \deltapos{15.16}} \\
      \ourrow
      \quad + GradFix & 20 & \redxmark
      & 66.76 & 75.08 & 68.40 & 73.54 & 61.48 & 85.85 & 90.60 & 73.19 & \textbf{74.36 \deltapos{14.05}} \\
    \bottomrule
  \end{tabular}}
  \vspace{1em}
\end{table*}

\tit{Language.} To assess whether task-vector transport generalizes beyond vision models, we evaluate \methodname in a language setting using an encoder--decoder Transformer~\citep{raffel2020exploring}. We consider \texttt{T5-3B} as the source model $A$ and \texttt{T5-Large} as the target model $B$. Both models are treated as text encoders ($\theta$) followed by task-specific classification heads ($\omega$). Our protocol is designed to isolate the effect of the transported \emph{encoder} representations. We first fine-tune model $A$ on each downstream task, computing a task-specific encoder update $\tau_A$ for $\theta_A$. We then transport this encoder update to model $B$ using \methodname, yielding the updated encoder $\theta_B + \tau_B$.

To evaluate the quality of the learned representations, we adopt a linear probing protocol \citep{alain2017understanding}. In all experiments, encoder parameters are kept strictly frozen, and a newly initialized linear classification head is trained from scratch. We consider three fixed encoders: the source encoder $\theta_A$, the target encoder $\theta_B$, and the target encoder augmented with the transported task update, $\theta_B + \tau_{B}$ (\methodname). All heads are trained under identical optimization settings, ensuring that performance differences reflect solely the quality of the underlying representations rather than any additional encoder fine-tuning.

We evaluate this setup on five closed-vocabulary natural language inference tasks from the GLUE benchmark \citep{wang2018glue}. We vary the number of batches $\mathcal{B}$ used to estimate the Procrustes maps and to train the classifier, using the same data budget for all methods. A detailed description of datasets and preprocessing is provided in \cref{app:text}. This evaluation protocol allows us to directly measure how much task-relevant structure is transferred by \methodname into the representation space of the target encoder, without interference from further encoder fine-tuning. By constraining adaptation to a linear classifier, we test whether the transported update reorganizes the target representation space in a task-aware manner.

As shown in \cref{tab:t5_lp}, transporting the encoder update from \texttt{T5-3B} to \texttt{T5-Large} consistently improves linear probing performance across all evaluated tasks and data regimes. While a gap relative to fully fine-tuned models remains, \methodname recovers a substantial fraction of the task-specific signal using only forward passes and closed-form alignment. Performance improves monotonically as the number of alignment batches $\mathcal{B}$ increases, indicating that richer activation statistics lead to more accurate alignment of the linguistic representation spaces.

Notably, the gains are obtained while keeping the target encoder completely frozen, suggesting that the transported task vector reorganizes the target representation space in a task-aware manner rather than merely transferring superficial classifier-level statistics. These results demonstrate that functional task-vector transport extends beyond vision models and applies effectively to encoder--decoder language architectures despite their additional structural complexity.

To evaluate \methodname on a non-classification task, we transport a summarization task update from \texttt{Flan-T5} to \texttt{T5-v1.1}. Using the ROUGE-L~\citep{lin-2004-rouge} metric, we observe a 36.5\% relative improvement over the zero-shot target model, demonstrating that our activation-alignment mechanism extends beyond classification to a generative language task. Additional details are reported in \cref{sec:t5_suppl}.

\begin{table*}[t]
\caption{\textbf{Depth and Width Scaling.} Task-vector transport from a ViT-B/16 model to a ViT-L/14 model, both pre-trained on \texttt{DataComp-XL}. This setting involves simultaneous depth and width mismatches and uses a naive layer-interpolation heuristic to align Transformer depth before transport. Results are reported for $\mathcal{B}=10$ alignment batches. $\Delta$Acc denotes the average accuracy improvement over the zero-shot baseline of model $B$. \methodname represents $\theta_B + \tau_B$.}\label{tab:diff_arch_vitb_vitl}
  \centering
  \resizebox{\linewidth}{!}{
  \begin{tabular}{ll *{9}{c}}
    \toprule
    \textbf{Model} & \multicolumn{1}{c}{$\mathcal{B}$} &
    \textbf{EUROSAT} & \textbf{SVHN} & \textbf{GTSRB} & \textbf{RESISC45} & \textbf{DTD} &
    \textbf{CARS} & \textbf{MNIST} & \textbf{SUN397} & \textbf{AVG ($\Delta$Acc)} \\
    \midrule
    $\theta_A$ \textit{interpolated} & \multicolumn{1}{c}{--}
  & 16.04 & 18.37 & 6.44 & 4.78 & 3.99 & 0.72 & 16.58 & 1.07 & 8.50 \deltaneg{65.13} \\
    \midrule
    $\theta_{B}$ \textit{zero-shot} & \multicolumn{1}{c}{--}
      & 68.40 & 67.44 & 58.96 & 71.27 & 66.81 & 93.01 & 86.84 & 76.34 & 73.63 \deltazero \\
    $\theta_{A}$ \textit{fine-tune} & \multicolumn{1}{c}{--}
      & 98.66 & 97.70 & 98.77 & 95.40 & 83.29 & 92.43 & 99.64 & 79.91 & 93.23 \deltapos{19.60} \\
    $\theta_{B}$ \textit{fine-tune} & \multicolumn{1}{c}{--}
      & 99.00 & 98.22 & 98.42 & 96.73 & 86.88 & 94.80 & 99.76 & 82.94 & 94.59 \deltapos{20.96} \\
    \midrule
    \ourrow
    \methodname & 10
      & 76.78 & 78.34 & 73.59 & 75.02 & 72.71 & 93.10 & 95.18 & 77.16 & \textbf{80.24 \deltapos{6.61}} \\
    \bottomrule
  \end{tabular}}
  \vspace{.5em}
\end{table*}

\subsection{Transfer on Identical Architectures}\label{sec:transport_same_arch}
We further evaluate \methodname in a controlled setting where models share an identical ViT-B/16 architecture but differ only in their pre-training distributions, mirroring the protocol from~\citet{rinaldi2026gradient}. Specifically, we transport task updates from a source model $A$ pre-trained on \texttt{DataComp-XL} to a target model $B$ pre-trained on \texttt{LAION-2B}. This configuration enables a direct comparison with TransFusion~\citep{rinaldi2025update} and GradFix. Following prior work, we adopt a standard $\mathcal{K}$-shot protocol, where $\mathcal{K}$ denotes the number of examples per class used for alignment or optimization.

Results in~\cref{tab:same_arch} show that our method consistently outperforms TransFusion across all shot regimes and remains highly competitive with GradFix under the same conditions. While GradFix relies on repeated backward passes and iterative optimization, \methodname achieves comparable or superior performance using only forward activations and a closed-form transport rule. Moreover, performance scales favorably with increasing supervision, matching or exceeding GradFix in higher-shot regimes. These results show that functional matching via orthogonal alignment provides an effective alternative to gradient-based task-vector transport even when models share the same architecture, demonstrating that the proposed formulation is not limited to heterogeneous-width settings.

\subsection{Transfer across Depth and Width}\label{sec:transport_diff_arch}
We finally assess structural generalization by transporting task updates between models that differ in both depth and width. Specifically, we transfer updates from a ViT-B/16 source model $A$ to a ViT-L/14 target model $B$, both pre-trained on \texttt{DataComp-XL}. This setting is particularly challenging since it requires reconciling mismatches in both representation dimensionality and transformer depth.

To extend \methodname beyond equal-depth settings, we adopt a simple interpolation-based heuristic to induce a coarse layer correspondence. Starting from a ViT-B model with $12$ transformer blocks, we construct a depth-expanded model by inserting one additional block after each original block, yielding a $24$-layer architecture compatible with ViT-L. Each inserted block $B_i'$ is obtained by linearly interpolating the parameters of adjacent blocks,
$\theta_{B_i'} = 0.5\,\theta_{B_i} + 0.5\,\theta_{B_{i+1}}$,
while the final inserted block is a copy of the last transformer block. The interpolation is applied to all transformer parameters, including LayerNorms, attention projections, MLP layers, and LayerScale coefficients. The same interpolation procedure is applied independently to both the base model $A$ and its fine-tuned counterpart $A^{ft}$, and the enlarged task update is computed from the resulting interpolated models. We then apply \methodname to handle the remaining width mismatch between the two architectures.

We emphasize that this depth alignment is heuristic and intentionally naive. Indeed, interpolation is highly destructive: the source model drops to $8.50\%$ zero-shot accuracy and $17.88\%$ fine-tuned accuracy after interpolation alone. Consequently, interpolation does not preserve the source predictor and serves only to induce a depth correspondence.

Nevertheless, as shown in~\cref{tab:diff_arch_vitb_vitl}, \methodname still enables effective task-vector transport on top of this degraded initialization, consistently improving over the target ViT-L zero-shot baseline. These results suggest that, despite severe parameter-space distortion, sufficient layerwise structure remains for activation-space alignment to recover transferable task directions. A more principled treatment of depth alignment remains an important direction for future work.

\subsection{Sensitivity to Update Scaling}
Standard task-vector transport and merging methods~\citep{marczak2025task,ilharco2023editing,tsv,yadav2023tiesmerging,rinaldi2026gradient} typically rely on a labeled validation set to select the optimal scalar interpolation coefficient $\alpha$ for integrating the transported update, applied as $\theta + \alpha\tau$. While this avoids gradient-based optimization on the target model, the reliance on target-task validation data partially relaxes the strict notion of training-free transfer. To evaluate the robustness of \methodname in a genuinely label-free regime, we analyze its performance across a sweep of $\alpha$ values, with particular interest in the unscaled $\alpha=1.0$ setting, which eliminates the need for any downstream validation search.

\begin{figure}[t]
\vspace{-0.5em}
    \centering
    \includegraphics[width=\linewidth]{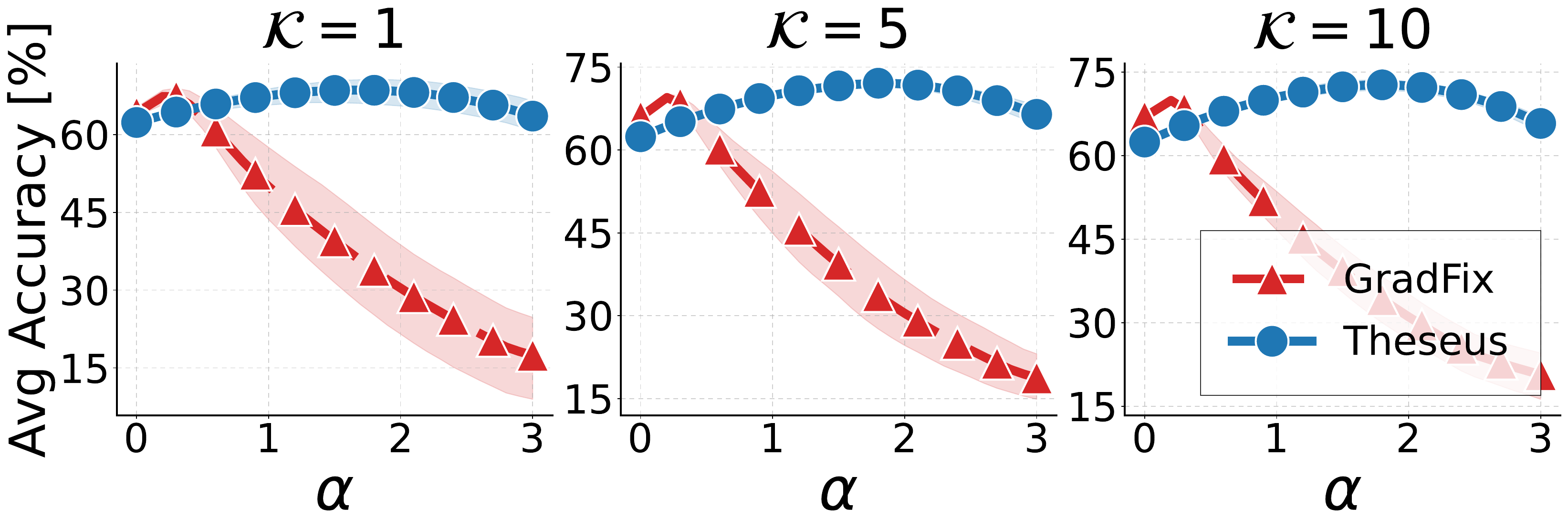}
    \caption{\textbf{Robustness to update scaling ($\alpha$).} \methodname exhibits high stability across different fixed $\alpha$ values, whereas GradFix experiences severe fluctuations in accuracy.}
    \label{fig:alpha_sweep}
    \vspace{-1em}
\end{figure}

As depicted in \cref{fig:alpha_sweep}, \methodname remains robust to scaling adjustments, maintaining a stable plateau of high accuracy across a wide range of $\alpha$ values and support set sizes ($\mathcal{K}$). In contrast, gradient-based baselines such as GradFix exhibit extreme sensitivity to this scaling coefficient, peaking sharply at low $\alpha$ values before decaying precipitously. \methodname scores $71.21/73.32/73.33$ for $\alpha=1/1.5/2$, while GradFix drops from $54.54$ to $44.64/36.12$.

This stability highlights a theoretical advantage of our formulation. By explicitly preserving the Frobenius norm and the geometric structure of the original update during transport (\cref{eq:norm_preservation}), the unscaled vector ($\alpha=1.0$) inherently retains its calibrated functional effect. Consequently, \methodname is uniquely suited for strictly label-free deployments where gradient-based alternatives suffer severe degradation.

\subsection{\methodname as Warm-Start for Fine-Tuning}
\label{sec:convergence_analysis}
\begin{figure}[t]
    \centering
    \includegraphics[width=\linewidth]{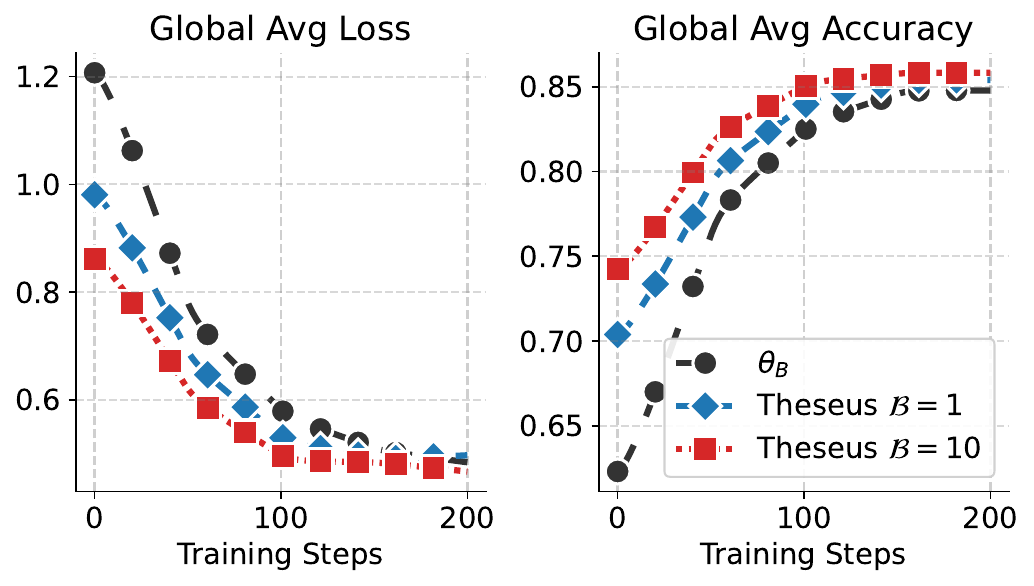}
    \caption{\textbf{Early convergence during downstream fine-tuning.} Average validation loss and accuracy on 8-Vision for standard fine-tuning from the baseline $\theta_B$ and from initialization with \methodname.}\label{fig:ft_from_tv}
\end{figure}

Finally, we investigate whether \methodname can improve downstream fine-tuning efficiency by providing a strong initialization for subsequent optimization. Specifically, we use the transported task vector to initialize the target model $\theta_B$ prior to standard supervised fine-tuning. \cref{fig:ft_from_tv} reports the average loss and accuracy over eight vision classification tasks during the first training steps, capturing the early convergence before all models reach similar minima. Models with \methodname start with a better initialization, exhibiting lower initial loss and higher accuracy than the standard pre-trained baseline. This warm-start effect is particularly evident in the convergence dynamics: the $10$-batch alignment variant achieves after only $50$ steps what the baseline $\theta_B$ reaches after $150$ steps. 

Overall, these results indicate that \methodname not only enables effective zero-shot adaptation but also serves as a practical initialization strategy that reduces the computational cost of supervised fine-tuning.

\subsection{Activation Space Alignment Analysis}
\label{sec:cosine_similarity}
To evaluate the effectiveness of the alignment, we measure the cosine similarity between intermediate activations from model $A$ ($H_A$) and model $B$ ($H_B$) before and after applying the transformation $T$. To ensure an out-of-sample evaluation, the alignment maps are estimated using calibration batches, while cosine similarity is measured on a disjoint validation set.

\begin{figure}[t]
    \centering
    \includegraphics[width=\linewidth]{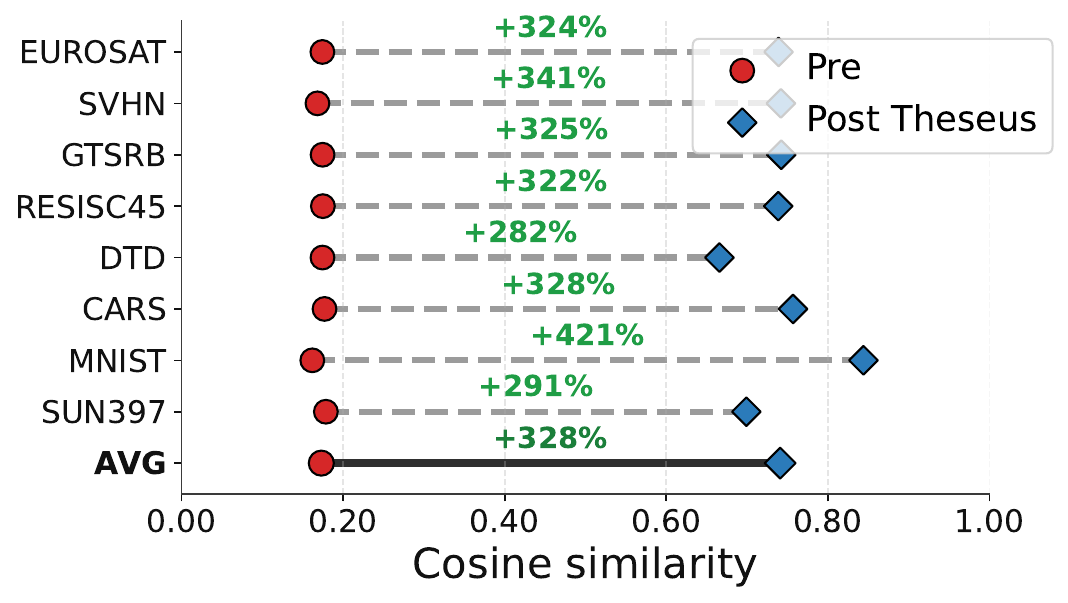}
    \caption{\textbf{Activation cosine similarity.} Cosine similarity between intermediate activations from the source and target models before (Pre) and after Procrustes alignment (Post \methodname).}
    \label{fig:activation_cosine}
\end{figure}

As shown in \cref{fig:activation_cosine}, the pre-trained models and their mismatched widths initially produce poorly aligned representation spaces, resulting in low baseline similarity (``Pre''). Applying the Procrustes transformations to project the source activations into the target representation space (``Post \methodname'') improves the correspondence between activations. On 8-Vision, \methodname yields an average relative increase in cosine similarity of $+328\%$.

These results provide strong empirical support for our central hypothesis: although the raw parameter spaces of heterogeneous models are not directly compatible, their representations remain organized around a shared latent functional subspace. By identifying and aligning this subspace, \methodname establishes the geometric basis necessary for accurate functional transport of task-specific updates.

\section{Conclusions}\label{sec:conclusions}
In this work, we introduced \textsc{Theseus}, a training-free framework for transporting task-specific updates across heterogeneous pre-trained models. By formulating task-vector transfer as a functional matching problem and conditioning it through orthogonal Procrustes alignment, \methodname yields a simple closed-form transport rule that preserves the geometric structure of task updates while naturally accommodating mismatches in representation dimensionality.

Across a wide range of experiments, we showed that \methodname enables transfer of task knowledge across models of different widths, pre-training distributions, and architectural depths. The method consistently outperforms naive baselines and direct pseudo-inverse approaches while remaining competitive with gradient-based alternatives when the architectures match, despite requiring only forward passes.

We also showed that transported updates provide a strong initialization for downstream fine-tuning, accelerating convergence and improving adaptation efficiency. More broadly, our results support a functional view of task identity, in which task-specific knowledge is characterized by its effect on representations. This suggests that task updates can remain transferable even across evolving architectures, widths, and pre-training distributions, provided that their underlying representation geometry is properly aligned. We believe this perspective opens promising directions for modular model adaptation, continual reuse of task updates, and scalable transfer across increasingly heterogeneous models.

\clearpage

\section*{Acknowledgments}

We acknowledge the CINECA award under the ISCRA initiative, for the availability of high-performance computing resources and support. The research activities of Angelo Porrello have been supported by the Department of Engineering ``Enzo Ferrari'' through the program FAR2025DIP (CUP E93C25000370005).  The work was partially funded by: Villanova, a project financed by IPICEI CIS, Prog. n. SA. 102519 - CUP B29J24000850005.

\section*{Impact Statement}

This work introduces \textsc{Theseus}, a training-free approach for transferring task-specific updates across pre-trained models by matching their functional effect on internal representations. By enabling reuse of task adaptations without additional optimization, our method has the potential to reduce computational costs, energy consumption, and environmental impact associated with repeatedly fine-tuning large models. This can facilitate more efficient deployment of machine learning systems, especially in resource-constrained settings or when adapting models across evolving architectures.

The proposed framework may also support more modular and sustainable model development practices, allowing practitioners to share and reuse task knowledge across model variants without access to original training data. However, as with any method that enables model adaptation, care should be taken to ensure that transferred task updates do not propagate unintended biases or harmful behaviors learned during fine-tuning. Our method preserves functional behavior by design, and thus inherits both the strengths and limitations of the source task updates.

We do not foresee direct misuse scenarios unique to \methodname beyond those already present in standard fine-tuning and model reuse pipelines. Nonetheless, responsible deployment should include appropriate evaluation and monitoring when transferring task updates across domains or applications. We hope this work encourages further research into efficient, transparent, and accountable mechanisms for model adaptation.

\bibliography{main}
\bibliographystyle{icml2026}

\clearpage
\appendix
%\onecolumn
\section{Closed-form Transport Rule}
\label{app:closed_form_derivation}

Starting from the surrogate objective introduced in the main text,
\begin{equation}
\label{eq:app_surrogate}
\min_{\tau_B}
\left\|
H_{\mathrm{in},A}
\left(
\tau_A^{\top}
-
T_{\mathrm{in}}\tau_B^{\top}T_{\mathrm{out}}^{\top}
\right)
H_{\mathrm{out},A}^{\top}
\right\|_F^2,
\end{equation}
define the residual operator
\begin{equation}
\label{eq:app_error}
E :=
\tau_A^{\top}
-
T_{\mathrm{in}}\tau_B^{\top}T_{\mathrm{out}}^{\top}.
\end{equation}
The objective can then be rewritten as
\begin{equation}
\label{eq:app_objective_E}
\min_{\tau_B}
\;
\|
H_{\mathrm{in},A} E H_{\mathrm{out},A}^{\top}
\|_F^2.
\end{equation}

Under the aligned-subspace approximation induced by Procrustes alignment, the transport objective is minimized when the residual term vanishes, \ie,
\begin{equation}
E = 0.
\end{equation}
Substituting~\cref{eq:app_error} gives
\begin{equation}
T_{\mathrm{in}}
\tau_B^{\top}
T_{\mathrm{out}}^{\top}
=
\tau_A^{\top}.
\end{equation}
Multiplying by $T_{\mathrm{in}}^\top$ on the left and $T_{\mathrm{out}}$ on the right, and using the row-orthonormality of the Procrustes maps, yields
\begin{equation}
\tau_B^{\top}
=
T_{\mathrm{in}}^\top
\tau_A^{\top}
T_{\mathrm{out}},
\end{equation}
which gives the transport rule
\begin{equation}
\label{eq:app_transport_rule}
\tau_B
=
T_{\mathrm{out}}
\tau_A
T_{\mathrm{in}}^{\top}.
\end{equation}

This derivation shows that the transported update corresponds to rotating the source task operator through the aligned representation subspaces identified by orthogonal Procrustes analysis.

\section{Orthogonality, Pseudo-Inverse Transport, and Alignment Objectives}
\label{app:pinv_relation}

\subsection{Underdetermined Nature of Least-Squares Alignment}

A natural approach to solving the transport objective~\cref{eq:transport_obj} is to derive a least-squares solution directly from the observed activations. Consider the alignment problem
\begin{equation}
\label{eq:ls_alignment_appendix}
\min_T \|H_A T - H_B\|_F^2,
\end{equation}
where $H_A \in \mathbb{R}^{N \times d_A}$ and $H_B \in \mathbb{R}^{N \times d_B}$ denote sampled activations from the source and target models.

When $H_A$ is rank-deficient, the problem admits infinitely many equivalent minimizers. Indeed, if $T^\star$ is a solution, then for any perturbation $\Delta$ satisfying
\begin{equation}
H_A \Delta = 0,
\end{equation}
we also have
\begin{equation}
H_A(T^\star + \Delta)=H_A T^\star.
\end{equation}

Thus, unconstrained least-squares alignment leaves null-space directions unconstrained. These directions can arbitrarily amplify poorly conditioned modes without affecting the reconstruction objective, leading to unstable transport operators and severe norm distortion.

This ambiguity becomes particularly problematic when transporting task updates across models with mismatched widths, since the activation covariances are often low-rank or only partially observed.

\subsection{Orthogonal Procrustes as a Regularized Alignment Objective}

To regularize the alignment problem, \methodname{} restricts the transport maps to have orthonormal rows:
\begin{equation}
TT^\top = I.
\end{equation}

Expanding the Frobenius objective in Eq.~\eqref{eq:ls_alignment_appendix},
\begin{align*}
\|H_A T - H_B\|_F^2
&=
\mathrm{tr}\!\left((H_A T - H_B)^\top(H_A T - H_B)\right)
\\
&=
\mathrm{tr}(T^\top H_A^\top H_A T)
-
2\mathrm{tr}(T^\top H_A^\top H_B)
\nonumber\\
&\quad+
\mathrm{tr}(H_B^\top H_B).
\end{align*}

Under the orthogonality constraint,
\begin{align}
\mathrm{tr}(T^\top H_A^\top H_A T)
&=
\mathrm{tr}(H_A T T^\top H_A^\top)
\\
&=
\mathrm{tr}(H_A H_A^\top)
=
\|H_A\|_F^2,
\end{align}
which is constant with respect to $T$. Therefore, the constrained problem reduces to
\begin{equation}
\max_{T:\,TT^\top=I}
\mathrm{tr}(T^\top C),
\qquad
C = H_A^\top H_B.
\end{equation}

Let
\begin{equation}
C = U \Sigma V^\top
\end{equation}
be the singular value decomposition of the cross-covariance matrix. The optimal orthogonal alignment is then given by
\begin{equation}
T^\star = UV^\top.
\end{equation}

Therefore, orthogonal Procrustes is not introduced as a heuristic assumption that the two representation spaces are exactly isometric. Rather, it emerges naturally as the closed-form optimizer of a regularized alignment problem that removes scaling ambiguities while preserving geometric structure.

\subsection{Relation to Pseudo-Inverse Transport}
\label{sec:pinv_relation}
A direct least-squares solution of the transport objective~\cref{eq:transport_obj} yields
\begin{equation}
\label{eq:app_pinv_solution}
\tau_B^{\top}
=
\left(H_{\mathrm{in},B}^{\top} H_{\mathrm{in},B}\right)^{\dagger}
H_{\mathrm{in},B}^{\top}
G_A
H_{\mathrm{out},B}
\left(H_{\mathrm{out},B}^{\top} H_{\mathrm{out},B}\right)^{\dagger},
\end{equation}
where $(\cdot)^{\dagger}$ denotes the Moore--Penrose pseudo-inverse.

However, this formulation is often numerically unstable in practice. When the two models differ in width, the activation Gram matrices
$H_{\mathrm{in},B}^{\top} H_{\mathrm{in},B}$ and
$H_{\mathrm{out},B}^{\top} H_{\mathrm{out},B}$
are frequently ill-conditioned or rank-deficient, especially when estimated from a finite number of samples.

Our Procrustes-based conditioning step can be interpreted as an orthogonal change of basis that aligns the dominant shared subspaces of the two representation spaces. In the aligned coordinates, the transport becomes well-conditioned and reduces to the closed-form transport rule derived in \cref{app:closed_form_derivation}.

Empirically, this distinction is critical: unconstrained least-squares transport suffers from severe norm amplification and unstable performance, whereas orthogonal alignment yields stable transport with near-perfect norm preservation.

\section{Experimental setting}\label{sec:exp_setting}

\tit{Models.}
In the vision experiments, we employ CLIP models \citep{radford2021learning} with architectures and weights obtained from OpenCLIP \citep{cherti2023reproducible}. We consider three distinct architectural configurations:
(i) \textbf{Width scaling}: transferring from a standard ViT-B/16 pre-trained on \texttt{LAION-2B} to a wider ViT-B/16+ variant pre-trained on \texttt{LAION-400M};
(ii) \textbf{Same architecture}: transferring between identical ViT-B/16 models pre-trained on different distributions, specifically \texttt{DataComp-XL} and \texttt{LAION-2B};
(iii) \textbf{Width and depth scaling}: transferring across both depth and width from a ViT-B/16 to a ViT-L/14, both pre-trained on \texttt{DataComp-XL}.
Unless otherwise specified, model $A$ denotes the smaller or source model and model $B$ the larger or target model. For the language experiments, we utilize T5 encoder--decoder models \citep{raffel2020exploring} of varying scales, specifically transferring from \texttt{T5-3B} to \texttt{T5-Large}. Code is implemented in the merge-and-rebase framework~\cite{panariello2026merge_and_rebase}.

% In the vision experiments, we consider CLIP models \citep{radford2021learning} based on the ViT-B architecture with varying widths and pre-training datasets, obtained from Open-CLIP \citep{cherti2023reproducible}. Unless otherwise specified, model $A$ denotes the narrower model and model $B$ the wider one. In the language experiments, we consider encoder-decoder Transformer T5 \citep{raffel2020exploring} models with differing parameter counts.

\tit{Task Updates.}
The task-specific updates $\tau_A$ are derived by fine-tuning the base model $A$ on the target downstream task via standard supervised objectives. Following the optimization protocol established by \citet{ilharco2023editing}, we fine-tune for $2000$ iterations with a batch size of $128$. We employ the AdamW optimizer \citep{loshchilov2018decoupled} with a peak learning rate of $1 \times 10^{-5}$, a weight decay of $0.1$, and a cosine annealing schedule preceded by
$200$ warm-up steps. In accordance with \citet{cherti2023reproducible}, the text encoder backbone remains frozen throughout the training process.
We then transport $\tau_A$ to model $B$ using the proposed method, without performing any additional optimization or gradient-based updates on model $B$.

\tit{Evaluation.}
For vision tasks, we evaluate classification accuracy on standard benchmarks following the CLIP evaluation protocol. For language tasks, we report accuracy on each dataset. All results are averaged across tasks unless otherwise specified.

\section{Computational Complexity Analysis}
\label{sec:comp_efficiency_suppl}

We compare the computational and memory complexity of \methodname against gradient-based transport methods such as GradFix as the target width $d_B$ increases. Unlike optimization-based approaches, \methodname performs a one-shot closed-form transport using only forward activations and covariance accumulation, avoiding repeated backward passes and optimizer-state storage.

% \tit{Memory Complexity.}
% Gradient-based approaches must retain activations for backpropagation together with gradients, optimizer states, and intermediate tensors, leading to memory usage that grows rapidly with model width. In contrast, \methodname accumulates cross-covariance statistics in a streaming fashion and does not store full activation tensors across batches.

% For a representative layer mapping $d_A = 768 \rightarrow d_B = 896$, \methodname stores only the covariance statistics corresponding to \num{1382912} scalars ($\sim$5.28 MiB in fp32) per layer. Importantly, this memory cost depends only on the layer widths and remains independent of the number of alignment batches $\mathcal{B}$.

\tit{Memory Complexity.}
Gradient-based approaches must retain activations for backpropagation together with gradients, optimizer states, and intermediate tensors, leading to memory usage that grows rapidly with model width. In contrast, \textsc{Theseus} accumulates cross-covariance statistics in a streaming fashion and does not store full activation tensors across batches. 

Specifically, during the forward pass of each calibration mini-batch, the algorithm extracts the source and target activation tensors for each layer $l$ (denoted as $H_A^l$ and $H_B^l$). Instead of retaining these large tensors in memory, the method immediately computes their uncentered cross-product ($(H_A^l)^{\top}H_B^l$) alongside the column-wise sums of the activations ($\sum H_A^l$ and $\sum H_B^l$). These lower-dimensional batch-level statistics are added to running accumulators, and the original, memory-intensive activation tensors are detached and discarded. 

Once all $\mathcal{B}$ calibration batches have been processed, the final centered cross-covariance matrix is computed algebraically on demand. The global means $\mu_A$ and $\mu_B$ are derived by dividing the accumulated sums by the total token count $M$. The exact centered covariance is then constructed via the identity $C^l = (H_A^l)^{\top}H_B^l - M(\mu_A \otimes \mu_B)$. Because the raw tokens are never concatenated or stored across batches, the maximum memory footprint overhead is strictly bounded by the $\mathcal{O}(d_A \times d_B)$ size of the covariance accumulators, making the memory complexity entirely independent of the total sequence length or the number of calibration batches used.

\tit{Runtime Complexity.}
Using a fixed alignment set of $M=512$ tokens, \methodname incurs a one-time alignment cost consisting of activation collection, covariance accumulation, and a single SVD per layer. Gradient-based approaches may be slightly faster at very small widths, but their cost scales more aggressively due to repeated backward passes and iterative optimization.

\begin{figure}[t]
    \centering
    \includegraphics[width=\linewidth]{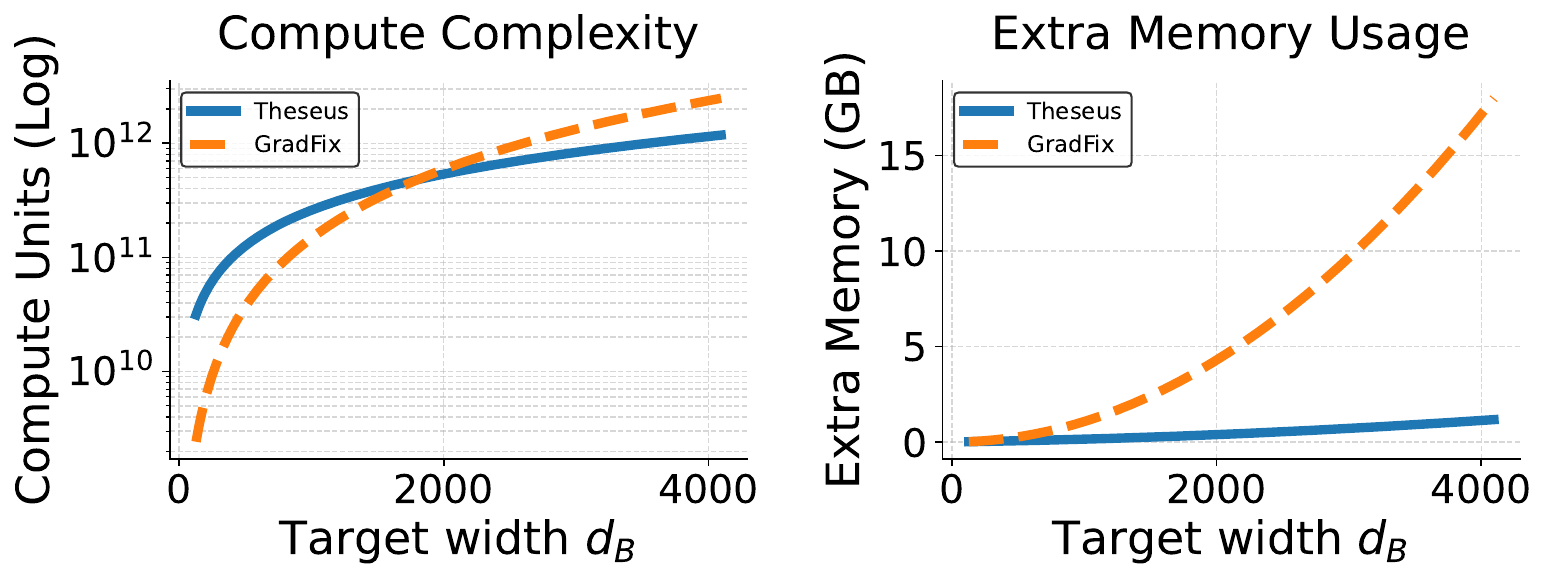}
    \caption{\textbf{Computational and memory scaling.} Runtime (left) and additional memory usage (right) as a function of the target model width $d_B$ (with source width $d_A=768$). While gradient-based fine-tuning methods such as GradFix exhibit steep growth due to backpropagation and activation storage, \methodname scales substantially better and maintains a nearly flat memory profile thanks to its streaming covariance formulation.}
    \label{fig:cost_scaling}
\end{figure}

As shown in \cref{fig:cost_scaling}, \methodname exhibits substantially more favorable scaling behavior as target width increases, making it particularly attractive for large-scale target architectures.

\section{Additional Experiments}\label{sec:additional_experiments}
\subsection{Sequence Length Alignment Strategies.}
\label{sec:interpolation_strategies}
\begin{table*}[t]
  \caption{\textbf{Ablation on Interpolation Modes.} Performance across datasets for different values of $\mathcal{B}$ and interpolation strategies. Results are reported as accuracy (\%).}
  \label{tab:ablation_interpolation}
    \centering
  \resizebox{\linewidth}{!}{
  \begin{tabular}{lc c cccccccc c}
    \toprule
    \textbf{Model} &  $\mathcal{B}$ & \textbf{Interpolation} & \textbf{EUROSAT} & \textbf{SVHN} & \textbf{GTSRB} & \textbf{RESISC45} & \textbf{DTD} & \textbf{CARS} & \textbf{MNIST} & \textbf{SUN397} & \textbf{AVG ($\Delta$Acc)} \\
    \midrule
    $\theta_{B}$ \textit{zero-shot} & -- & -- & 56.69 & 62.66 & 55.46 & 68.51 & 58.19 & 88.81 & 76.47 & 72.42 & 67.40 \deltazero \\
    $\theta_{B}$ \textit{fine-tune} & -- & -- & 98.66 & 97.70 & 98.77 & 95.40 & 83.29 & 92.43 & 99.64 & 79.91 & 93.23 \deltapos{25.83} \\
    \midrule
    
    \multirow{3}{*}{\methodname} & \multirow{3}{*}{1} 
      & mean          & 59.60 & 63.21 & 55.47 & 68.45 & 58.33 & 88.89 & 79.00 & 72.53 & 68.19 \deltapos{0.79} \\
    & & interpolate   & 57.96 & 62.91 & 56.88 & 68.75 & 58.85 & 88.92 & 79.33 & 72.49 & 68.23 \deltapos{0.83} \\
    \ourrow
    \cellcolor{white}
    & \cellcolor{white} & interpolate2d & 66.31 & 76.55 & 63.57 & 70.33 & 60.00 & 88.86 & 88.14 & 72.67 & \textbf{73.30 \deltapos{5.90}} \\
    \midrule

    \multirow{3}{*}{\methodname} & \multirow{3}{*}{2} 
      & mean          & 58.31 & 62.96 & 56.72 & 68.73 & 58.39 & 88.83 & 79.10 & 72.47 & 68.44 \deltapos{1.04} \\
    & & interpolate   & 62.94 & 63.33 & 57.13 & 69.03 & 58.97 & 88.98 & 78.66 & 72.64 & 68.96 \deltapos{1.56} \\
    \ourrow
    \cellcolor{white}
    & \cellcolor{white}& interpolate2d & 70.00 & 73.86 & 68.38 & 72.59 & 59.95 & 88.85 & 88.64 & 72.75 & \textbf{74.38 \deltapos{6.98}} \\
    \midrule

    \multirow{3}{*}{\methodname} & \multirow{3}{*}{5} 
      & mean          & 62.33 & 63.20 & 58.16 & 68.89 & 59.41 & 88.88 & 80.94 & 72.88 & 69.34 \deltapos{1.94} \\
    & & interpolate   & 62.31 & 64.25 & 58.79 & 69.71 & 59.81 & 89.08 & 82.27 & 72.77 & 69.88 \deltapos{2.48} \\
    \ourrow
    \cellcolor{white}
    & \cellcolor{white}& interpolate2d & 71.14 & 79.66 & 69.78 & 72.27 & 62.02 & 88.95 & 93.88 & 73.01 & \textbf{76.34 \deltapos{8.94}} \\
    \bottomrule
  \end{tabular}}
\end{table*}

To resolve sequence length mismatches between the source $\theta_A$ and target $\theta_B$, we evaluate three distinct alignment strategies. These strategies are necessary because different ViT configurations may encode images into different numbers of patches, resulting in distinct token sequence lengths $L^A$ and $L^B$.

The \textit{mean} strategy serves as a baseline by averaging the feature representations across the sequence dimension, effectively removing spatial granularity. The \textit{interpolation} strategy treats the token sequences as 1D signals and uses linear interpolation to resample the source sequence to the target length. Finally, \textit{interpolation2d} leverages the spatial inductive bias of the Vision Transformer by reshaping the 1D token sequence into a 2D grid that mirrors the original image patch layout. A bilinear interpolation is then applied to this grid to match the target resolution before flattening the result back into a sequence.
\begin{table*}[t]
  \caption{\textbf{Width Scaling (Narrow $\to$ Wide)} Cross-dataset performance ($A$: \texttt{laion2b} $\to$ $B$: \texttt{laion400m}). Results are reported for varying support set sizes $\mathcal{K}$}
  \label{tab:small_to_big_ds}
  \centering
  \resizebox{\linewidth}{!}{
  \begin{tabular}{l c c c c c c c c c c}
    \toprule
    \textbf{Model} & $\mathcal{K}$ & \textbf{EUROSAT} & \textbf{SVHN} & \textbf{GTSRB} & \textbf{RESISC45} & \textbf{DTD} & \textbf{CARS} & \textbf{MNIST} & \textbf{SUN397} & \textbf{AVG ($\Delta$Acc)} \\
    \midrule
    $\theta_{B}$ \textit{zero-shot}
      & -- & 50.92 & 39.23 & 49.63 & 64.53 & 55.48 & 84.53 & 57.06 & 68.67 & 58.76 \deltazero \\
    $\theta_{B}$ \textit{fine-tune}
      & -- & 98.96 & 91.08 & 98.63 & 92.59 & 77.81 & 87.65 & 99.63 & 76.76 & 90.39 \deltapos{31.63} \\
    $\theta_{A}$ \textit{fine-tune}
     & -- & 98.69 & 97.45 & 98.64 & 95.65 & 82.24 & 91.53 & 99.61 & 79.86 & 92.96 \deltapos{34.20} \\
    \midrule
    \ourrow
    \methodname
      & 1 & 60.79 & 55.73 & 57.39 & 67.48 & 57.34 & 85.02 & 61.67 & 70.76 & \textbf{64.52 \deltapos{5.76}} \\
    \ourrow
    \methodname
      & 2 & 61.46 & 56.07 & 59.03 & 68.01 & 58.56 & 85.06 & 68.19 & 70.81 & \textbf{65.90 \deltapos{7.14}} \\
    \ourrow
    \methodname
      & 5 & 64.58 & 55.20 & 59.05 & 69.18 & 59.25 & 85.17 & 77.57 & 70.93 & \textbf{67.62 \deltapos{8.86}} \\
    \ourrow
    \methodname
      & 10 & 65.51 & 56.86 & 59.72 & 69.48 & 60.48 & 85.16 & 76.26 & 71.28 & \textbf{68.09 \deltapos{9.33}} \\
    \ourrow
    \methodname
      & 20 & 66.83 & 60.33 & 58.25 & 70.03 & 60.53 & 85.51 & 78.38 & 72.12 & \textbf{69.00 \deltapos{10.24}} \\
    \bottomrule
  \end{tabular}}
\end{table*}
\begin{table*}[t]
  \caption{\textbf{Width Scaling (Wide $\to$ Narrow)} Cross-dataset performance ($A$: \texttt{laion400m} $\to$ $B$: \texttt{datacomp\_xl}). Results are reported for varying support set sizes $\mathcal{K}$}
  \label{tab:big_to_small_ds}
  \centering
  \resizebox{\linewidth}{!}{
  \begin{tabular}{lc *{9}{c}}
    \toprule
    \textbf{Model} & $\mathcal{K}$ &
    \textbf{EUROSAT} & \textbf{SVHN} & \textbf{GTSRB} & \textbf{RESISC45} & \textbf{DTD} &
    \textbf{CARS} & \textbf{MNIST} & \textbf{SUN397} & \textbf{AVG ($\Delta$Acc)} \\
    \midrule
    $\theta_{B}$ \textit{zero-shot} &
      & 56.69 & 62.66 & 55.46 & 68.51 & 58.19 & 88.81 & 76.47 & 72.42 & 67.40 \deltazero \\
    $\theta_{B}$ \textit{fine-tune} &
      & 98.66 & 97.70 & 98.77 & 95.40 & 83.29 & 92.43 & 99.64 & 79.91 & 93.23 \deltapos{25.83} \\
    \midrule
    \ourrow & 1
      & 62.95 & 73.56 & 66.99 & 71.03 & 60.16 & 88.79 & 83.31 & 73.18 & \textbf{72.50 \deltapos{5.10}} \\
    \ourrow & 2
      & 68.01 & 74.19 & 67.19 & 70.89 & 60.53 & 88.91 & 88.15 & 73.26 & \textbf{73.89 \deltapos{6.49}} \\
    \ourrow & 5
      & 68.35 & 75.92 & 71.20 & 72.68 & 62.13 & 88.92 & 92.25 & 73.22 & \textbf{75.58 \deltapos{8.18}} \\
    \ourrow & 10
      & 73.16 & 76.93 & 67.63 & 73.44 & 62.93 & 88.89 & 91.98 & 73.21 & \textbf{76.02 \deltapos{8.62}} \\
    \ourrow \multirow{-5}{*}{\methodname} & 20
      & 70.84 & 82.82 & 70.63 & 73.89 & 63.62 & 89.03 & 94.33 & 73.38 & \textbf{77.32 \deltapos{9.92}} \\
    \bottomrule
  \end{tabular}}
\end{table*}
\begin{table*}[t]
  \caption{\textbf{Width Scaling (Wide $\to$ Narrow)} Cross-dataset performance ($A$: \texttt{laion400m} $\to$ $B$: \texttt{datacomp\_xl}). Results for varying number of alignment batches $B$.}
  \label{tab:big_to_small_batch}
  \centering
  \resizebox{\linewidth}{!}{
  \begin{tabular}{lc *{9}{c}}
    \toprule
    \textbf{Model} & $\mathcal{B}$ &
    \textbf{EUROSAT} & \textbf{SVHN} & \textbf{GTSRB} & \textbf{RESISC45} & \textbf{DTD} &
    \textbf{CARS} & \textbf{MNIST} & \textbf{SUN397} & \textbf{AVG ($\Delta$Acc)} \\
    \midrule
    $\theta_{B}$ \textit{zero-shot} &
      & 56.69 & 62.66 & 55.46 & 68.51 & 58.19 & 88.81 & 76.47 & 72.42 & 67.40 \deltazero\\
    $\theta_{B}$ \textit{fine-tune} &
      & 98.66 & 97.70 & 98.77 & 95.40 & 83.29 & 92.43 & 99.64 & 79.91 & 93.23 \deltapos{25.83} \\
    \midrule
    \ourrow
    \methodname & 1
      & 66.31 & 76.55 & 63.57 & 70.33 & 60.00 & 88.86 & 88.14 & 72.67 & \textbf{73.30 \deltapos{5.90}} \\
      \ourrow
    \methodname & 2
      & 70.00 & 73.86 & 68.38 & 72.59 & 59.95 & 88.85 & 88.64 & 72.75 & \textbf{74.38 \deltapos{6.98}} \\
      \ourrow
    \methodname & 5
      & 71.14 & 79.66 & 69.78 & 72.27 & 62.02 & 88.95 & 93.88 & 73.01 & \textbf{76.34 \deltapos{8.94}} \\
      \ourrow
    \methodname & 10
      & 71.75 & 79.73 & 69.58 & 72.33 & 62.25 & 89.00 & 94.12 & 73.16 & \textbf{76.49 \deltapos{9.09}} \\
    \ourrow
    \methodname & 20
      & 72.36 & 80.33 & 68.58 & 74.01 & 62.61 & 88.97 & 94.30 & 73.24 & \textbf{76.80 \deltapos{9.40}} \\
    \bottomrule
  \end{tabular}}
\end{table*}

As shown in \cref{tab:ablation_interpolation}, the \textit{interpolation2d} strategy consistently yields the highest accuracy across all batch sizes, significantly outperforming both the $1D$ linear and mean-based approaches. This emphasizes the importance of preserving the spatial topology of the features: by maintaining the geometric correspondence between patches, the method more effectively transfers the task-specific knowledge distilled by the source model. Notably, these sequence-level adjustments are exclusive to the vision domain. In contrast, for the T5 experiments (\cref{tab:t5_lp,tab:t5}), no sequence interpolation is necessary. Because T5 natively supports variable-length text sequences, input lengths can be adjusted prior to activation computation to directly match source and target representations, allowing the transport process to be performed without additional preprocessing.

\subsection{Extended Results on Width Scaling.}
\label{sec:width_additional}
We provide a comprehensive evaluation of the ($A \to B$) transport direction under varying $\mathcal{K}$-shot settings in \cref{tab:small_to_big_ds}, illustrating how performance scales with the number of labeled samples per class. Complementary results for the reverse transport direction ($B \to A$) are detailed in \cref{tab:big_to_small_batch,tab:big_to_small_ds}, using both batch-based $\mathcal{B}$-shot and class-balanced $\mathcal{K}$-shot protocols. Collectively, these findings confirm that \methodname is robust across diverse sampling strategies and consistently yields improvements, even when the target model already exhibits a strong zero-shot baseline.

\subsection{Language.} 
\label{sec:t5_suppl}
\begin{table*}[t]
  \caption{\textbf{Width Scaling (Wide $\to$ Narrow).} Cross-dataset performance for transferred task-specific encoder representations ($A$: \texttt{T5-3B} $\to$ $B$: \texttt{T5-Large}). Results are reported for varying numbers of alignment batches $\mathcal{B}$.}
  \label{tab:t5}
  \centering

  \begin{tabular}{l c c c c c c c}
    \toprule
    \textbf{Model} & \textbf{$\mathcal{B}$} & \textbf{MNLI} & \textbf{QNLI} & \textbf{RTE} & \textbf{SCITAIL} & \textbf{SNLI} & \textbf{AVG ($\Delta$Acc)} \\
    \midrule
    $\theta_{B}$
      & -- & 32.23 & 48.05 & 48.38 & 50.77 & 32.44 & 42.37 \deltazero \\
    $\theta_{B}$ \textit{fine-tune}
      & -- & 89.80 & 94.27 & 83.59 & 95.86 & 91.83 & 91.09 \deltapos{48.72} \\
      \midrule
    \ourrow
    \methodname
       & 20 & 57.01 & 50.56 & 49.46 & 50.77  & 57.24 & \textbf{53.01 \deltapos{10.64}} \\
       \ourrow
    \methodname
       & 50 & 54.52 & 56.31 & 50.12 & 53.45 & 60.42 & \textbf{54.96 \deltapos{12.59}} \\
       \ourrow
    \methodname
      & 100 & 55.66 & 59.49 & 50.98 & 63.11 & 64.89 & \textbf{58.83 \deltapos{16.46}} \\
    \bottomrule
  \end{tabular}
\end{table*}

Following the experimental setting of \citet{rinaldi2025update}, we consider the model $\theta = \{\phi, \omega\}$ composed of a pre-trained feature extractor $\phi$ and a classification head $\omega$. We transport the task vector $\tau^\phi_A$ from model $A$ (\texttt{T5-3B}) to the target feature extractor $\phi_B$ (\texttt{T5-Large}). We then refrain from training a new classifier for the target model and instead reuse the source fine-tuned head $\omega_{ft}^A$. This allows us to evaluate whether the transported representation space aligns with the source model's classification logic without additional head optimization.
As baselines, we report the performance of the target backbone $\theta_B$ equipped with a randomly initialized classification head, alongside a fully fine-tuned version of $\theta_B$ where both the encoder and head are optimized for the target task.
As shown in Table~\ref{tab:t5}, the gains are modest but consistent across different values of $\mathcal{B}$, and \methodname outperforms the baseline $\theta_B$ in every setting.

\begin{table}[t]
\def\arraystretch{1.1}
  \caption{ROUGE-L scores for summarization transfer. Source: \texttt{Flan-T5} ($A$), target: \texttt{T5-v1.1} ($B$). \methodname represents $\theta_B + \tau_B$. Results are reported for varying numbers of alignment samples $\mathcal{D}$.}
  \label{tab:t5_sum}
  \centering
  \setlength{\tabcolsep}{4pt}
  \resizebox{\linewidth}{!}{
  \begin{tabular}{l c c c c c}
    \toprule
    \textbf{Configuration} & $\alpha$ & $\mathcal{D}=200$ & $\mathcal{D}=500$ & $\mathcal{D}=800$ & \textbf{Average} \\
    \midrule
    $\theta_{B}$ 
      & \multicolumn{1}{c}{--} 
      & \multicolumn{4}{c}{0.1053} 
       \\
      
    $\theta_{B}$ \textit{fine-tune} 
      & \multicolumn{1}{c}{--} 
      & \multicolumn{4}{c}{0.2821} 
       \\
      
    \midrule

    \ourrow
    \methodname & 0.5 & 0.1173 & 0.1201 & 0.1265 & 0.1213 \\

    \ourrow
    \methodname & 1.0 & 0.1278 & 0.1337 & 0.1437 & 0.1351 \\

    \ourrow
    \methodname & 2.0 & 0.1237 & 0.1189 & 0.1423 & 0.1283 \\

    \ourrow
    \methodname & 4.0 & 0.0876 & 0.0512 & 0.0879 & 0.0756 \\
    
    \bottomrule
  \end{tabular}}
\end{table}

\paragraph{Summarization Task.} As shown in \cref{tab:t5_sum}, \methodname can also be applied to the summarization task. \texttt{Flan-T5} was fine-tuned on the CNN/DailyMail dataset (\cref{app:text}) using the same hyperparameters as in the other experiments. The resulting task vector was then transported to \texttt{T5-v1.1}. Performance was evaluated using the ROUGE-L metric \citep{lin-2004-rouge}, which measures the longest common subsequence overlap between the generated and reference summaries.

\section{Dataset Details}
\subsection{Visual Datasets.}\label{sec:vision_exp}
We evaluate our method on a diverse suite of vision classification benchmarks, covering a wide range of domains from natural images to satellite imagery:
\begin{itemize}
    \item \textbf{EuroSAT:} A remote sensing dataset based on Sentinel-2 satellite imagery, comprising \num{27000} geo-referenced samples across 10 distinct classes \citep{eurosat}.
    \item \textbf{SVHN:} A digit recognition benchmark containing \num{73257} real-world images of house numbers across 10 classes extracted from Google Street View \citep{svhn}.
    \item \textbf{GTSRB:} The German Traffic Sign Recognition Benchmark, a standard for autonomous driving tasks featuring \num{51839} images across 43 traffic sign categories \citep{gtsrb}.
    \item \textbf{RESISC45:} A high-resolution remote sensing dataset containing \num{31500} images from Google Earth, categorized into 45 scene classes to evaluate fine-grained aerial classification \citep{cheng_remote_2017}.
    \item \textbf{DTD:} The Describable Textures Dataset, consisting of \num{5640} texture-centric images organized into 47 perceptual categories \citep{dtd}.
    \item \textbf{Stanford Cars:} A fine-grained classification dataset comprising \num{16185} images across 196 categories of cars, distinguished by make, model, and year \citep{cars}.
    \item \textbf{MNIST:} The Modified National Institute of Standards and Technology database, a foundational benchmark for handwritten digit recognition containing \num{70000} grayscale images across 10 classes \citep{MNIST}.
    \item \textbf{SUN397:} A large-scale scene understanding benchmark consisting of \num{108754} images covering 397 categories, ranging from indoor environments to diverse outdoor landscapes \citep{sun397}.

\end{itemize}
\subsection{Textual Datasets}\label{app:text}
\begin{itemize}
  \item \textbf{SNLI}: Stanford Natural Language Inference dataset, containing \num{570000} sentence pairs labeled for entailment, contradiction, or neutral \citep{bowman2015large}.
  \item \textbf{MNLI}: The Multi-Genre Natural Language Inference dataset contains \num{433000} sentence pairs annotated with textual entailment information across various genres \citep{mnli}.
  \item \textbf{RTE}: Recognizing Textual Entailment dataset, with \num{2490} examples for training, \num{277} for validation, and \num{3000} for testing, divided into two classes \citep{wang2018glue}.
  \item \textbf{QNLI}: The Question Natural Language Inference dataset contains \num{104743} training examples across two classes \citep{wang2018glue}.
  \item \textbf{SCITAIL}: A science entailment dataset built from science question answering, with \num{23596} training examples of two classes \citep{khot2018scitail}.
  \item \textbf{CNN/DailyMail}: An English-language dataset containing more than 300k news articles written by journalists at CNN and the Daily Mail, paired with summaries \citep{hermann2015teaching, nallapati2016abstractive}.
\end{itemize}

\section{Functional Transport vs. Feature Alignment}
\label{app:feature_vs_operator}

As observed in \cref{sec:transport_width} (\cref{tab:small_to_big_batch}), unconstrained pseudo-inverse baselines ($\tau_{\text{pinv}}$, $\tau_{\text{pinv-Tikh}}$) consistently underperform \methodname, particularly in low-shot regimes where the activation statistics become severely rank-deficient. In this section, we analyze the source of this instability and isolate the contributions of both our coupled functional objective (\cref{eq:transport_obj}) and the orthogonal constraints from standard latent-space feature alignment approaches \citep{moschella2023relative, maiorca2023latent}.

To do so, we ablate six alternative transport constructions across two fundamental axes: whether they optimize a coupled operator objective or decoupled feature maps, and whether they are unconstrained or orthogonally regularized (Tab.~\ref{tab:method_ablation}).

\subsection*{1. Coupled Unconstrained Transport}
These baselines share \methodname's functional matching objective, attempting to find a target update $\tau_B$ that directly minimizes the bilinear interaction mismatch:

\begin{itemize}
    \item \textbf{Pinv / Pinv-Tikh.}
    Direct least-squares solutions to the coupled objective. As derived in \cref{sec:pinv_relation}, the unconstrained Moore--Penrose pseudo-inverse yields:
    \begin{equation}
\begin{split}
    \tau_B^\top &= (H_{\mathrm{in},B}^\top H_{\mathrm{in},B})^\dagger H_{\mathrm{in},B}^\top (H_{\mathrm{in},A} \tau_A^\top H_{\mathrm{out},A}^\top) \\
    &\quad \times H_{\mathrm{out},B} (H_{\mathrm{out},B}^\top H_{\mathrm{out},B})^\dagger.
\end{split}
\end{equation}
    The Tikhonov variant adds ridge regularization ($\lambda I$) to the Gram matrices prior to inversion to improve numerical stability.
\end{itemize}

\subsection*{2. Decoupled Feature Alignment}
These baselines abandon the coupled operator objective. Instead, they estimate independent linear maps $R_{\mathrm{in}}$ and $R_{\mathrm{out}}$ to translate the input and output feature spaces, and form the update via $\tau_B = R_{\mathrm{out}} \tau_A R_{\mathrm{in}}^\top$.

\begin{itemize}
    \item \textbf{Linear.}
    Standard unconstrained least-squares alignment for each space independently:
    \begin{equation}
    R_{\mathrm{in}}^{\text{lin}}
    =
    (H_{\mathrm{in},A}^\top H_{\mathrm{in},A})^\dagger
    (H_{\mathrm{in},A}^\top H_{\mathrm{in},B}).
    \end{equation}

    \item \textbf{Ortho.}
    Orthogonal Procrustes alignment computed directly from the cross-covariance $H_{\mathrm{in},A}^\top H_{\mathrm{in},B} = U \Sigma V^\top$, yielding the orthogonal map $R_{\mathrm{in}}^{\text{ortho}} = UV^\top$.

    \item \textbf{Linear-Ortho / Pinv-Ortho.}
    Orthogonal projections of the unconstrained \textit{Linear} and \textit{Pinv} feature maps onto the nearest orthogonal matrix through SVD. This mirrors the standard pipeline in prior latent-space translation methods, where an unconstrained alignment is first estimated and subsequently orthogonalized.
\end{itemize}

\subsection*{3. Coupled Constrained Transport (\textsc{Theseus})}
Our proposed formulation combines the coupled functional objective with strict orthogonal constraints. This reduces to aligning the input and output cross-covariances ($C_{\mathrm{in}} = H_{\mathrm{in},A}^\top H_{\mathrm{in},B}$, $C_{\mathrm{out}} = H_{\mathrm{out},A}^\top H_{\mathrm{out},B}$), producing the closed-form Procrustes embeddings $T_{\mathrm{in}} = U_{\mathrm{in}}V_{\mathrm{in}}^\top$ and $T_{\mathrm{out}} = U_{\mathrm{out}}V_{\mathrm{out}}^\top$, and the transported update $\tau_B = T_{\mathrm{out}}\tau_A T_{\mathrm{in}}^\top$.

The results in \cref{tab:method_ablation} reveal the distinct failure modes of the baselines.

\textbf{The necessity of geometric constraints.} Unconstrained solutions to the coupled objective (\textit{Pinv}) exhibit severe norm amplification, particularly in low-shot settings where the activation covariances become poorly conditioned. Null-space directions are arbitrarily amplified, yielding massive norm distortion ($\|\tau_B\|_F \gg \|\tau_A\|_F$). While Tikhonov regularization (\textit{Pinv-Tikh}) partially alleviates this by damping high-gain directions, it introduces a sensitive hyperparameter and still distorts the update geometry.

\textbf{The necessity of the coupled objective.} Enforcing orthogonality (\textit{Ortho}, \textit{Linear-Ortho}, \textit{Pinv-Ortho}) resolves the instability. Because orthogonal maps preserve norms and angles, they eliminate scaling ambiguities and maintain $\|\tau_B\|_F \approx \|\tau_A\|_F$. However, despite this stabilization, all orthogonal feature-alignment baselines remain consistently below \methodname. This gap reveals that independent alignment of input and output representations is insufficient to guarantee preservation of the functional effect of the task update. There is no mathematical guarantee that composing independently optimal feature maps preserves the bilinear interaction structure encoded by $H_{\mathrm{in}}\tau^\top H_{\mathrm{out}}^\top$. Furthermore, for two-step variants like \textit{Pinv-Ortho} and \textit{Linear-Ortho}, computing an unconstrained mapping first and then projecting it onto the orthogonal group means orthogonalizing an already corrupted matrix: the initial inversion step amplifies noise in rank-deficient directions, locking in a suboptimal rotation.

In contrast, \methodname\ succeeds because it enforces geometric stability while deriving the transport rule directly from the coupled functional objective, effectively bypassing ill-conditioned intermediate inversions entirely.

\begin{table*}[t]
  \caption{\textbf{Alignment Method Ablation.} Accuracy and norm-preservation error ($\lVert\tau_B\rVert_F - \lVert\tau_A\rVert_F$) for different transport operators in the ViT-B/16 $\to$ ViT-B/16+ setting.}
  \label{tab:method_ablation}
  \centering
  \sisetup{detect-weight=true, mode=match}
  \begin{tabular}{c c S[table-format=2.2] S[table-format=-1.2e-2]}
    \toprule
    \textbf{$\mathcal{B}$} & \textbf{Method} & {\textbf{Accuracy (\%)}} & {\textbf{Norm Error}} \\
    \midrule

    \multirow{7}{*}{2} 
      & $\tau_{\text{pinv}}$       & 52.25 & 6.30e4 \\
      & $\tau_{\text{pinv-Tikh}}$  & 59.60 & 2.30e-4 \\
      & $\tau_{\text{ortho}}$      & 63.31 & 2.40e1 \\
      & $\tau_{\text{linear}}$     & 4.52  & 1.15e4 \\
      & $\tau_{\text{linear-ortho}}$ & 60.93 & -4.02e-9 \\
      & $\tau_{\text{pinv-ortho}}$ & 60.47 & 3.18e-11 \\
    \ourrow
      & \methodname                & \bfseries 66.60 & \bfseries 7.60e-10 \\

    \midrule

    \multirow{7}{*}{5} 
      & $\tau_{\text{pinv}}$       & 56.80 & 4.70e4 \\
      & $\tau_{\text{pinv-Tikh}}$  & 60.29 & 1.10e-4 \\
      & $\tau_{\text{ortho}}$      & 64.23 & 1.90e1 \\
      & $\tau_{\text{linear}}$     & 4.53  & 7.83e3 \\
      & $\tau_{\text{linear-ortho}}$ & 61.64 & 7.78e-9 \\
      & $\tau_{\text{pinv-ortho}}$ & 60.51 & 6.82e-11 \\
    \ourrow
      & \methodname                & \bfseries 66.64 & \bfseries 8.40e-10 \\

    \midrule

    \multirow{7}{*}{10} 
      & $\tau_{\text{pinv}}$       & 58.39 & 3.80e4 \\
      & $\tau_{\text{pinv-Tikh}}$  & 60.75 & 3.90e-7 \\
      & $\tau_{\text{ortho}}$      & 65.44 & 1.70e1 \\
      & $\tau_{\text{linear}}$     & 4.69  & 6.27e3 \\
      & $\tau_{\text{linear-ortho}}$ & 62.77 & 3.21e-9 \\
      & $\tau_{\text{pinv-ortho}}$ & 66.75 & -1.90e-10 \\
    \ourrow
      & \methodname                & \bfseries 68.35 & \bfseries 7.90e-10 \\

    \bottomrule
  \end{tabular}
\end{table*}

\end{document}